\documentclass[fleqn,10pt]{wlscirep}
\usepackage[utf8]{inputenc}
\usepackage[T1]{fontenc}
\usepackage{amsmath}
\usepackage{amsfonts}
\usepackage{multirow}
\usepackage{makecell}
\usepackage{booktabs}
\usepackage{hyperref}
\usepackage{array}
\usepackage{arydshln}
\usepackage{fancyhdr}
\usepackage{adjustbox}
\usepackage{tabularx}
\usepackage{algorithm}
\usepackage{algpseudocode}

\makeatletter
\DeclareRobustCommand\onedot{\futurelet\@let@token\@onedot}
\def\@onedot{\ifx\@let@token.\else.\null\fi\xspace}

\makeatother
\title{Supervised Pretraining for Material Property Prediction}

\author[1]{Chowdhury Mohammad Abid Rahman}
\author[2]{Aldo H. Romero}
\author[1,*]{Prashnna K. Gyawali}
\affil[1]{Lane Department of Computer Science and Electrical Engineering, West Virginia University}
\affil[2]{Department of Physics and Astronomy, West Virginia University}

\affil[*]{prashnna.gyawali@mail.wvu.edu}

\begin{abstract}
Accurate prediction of material properties facilitates the discovery of novel materials with tailored functionalities. Deep learning models have recently shown superior accuracy and flexibility in capturing structure-property relationships. However, these models often rely on supervised learning, which requires large, well-annotated datasets—an expensive and time-consuming process.
Self-supervised learning (SSL) offers a promising alternative by pretraining on large, unlabeled datasets to develop foundation models that can be fine-tuned for material property prediction. In this work, we propose supervised pretraining, where available class information serves as surrogate labels to guide learning, even when downstream tasks involve unrelated material properties. We evaluate this strategy on two state-of-the-art SSL models and introduce a novel framework for supervised pretraining. To further enhance representation learning, we propose a graph-based augmentation technique that injects noise to improve robustness without structurally deforming material graphs.
The resulting foundation models are fine-tuned for six challenging material property predictions, achieving significant performance gains over baselines, ranging from 2\% to 6.67\% improvement in mean absolute error (MAE)—and establishing a new benchmark in material property prediction. This study represents the first exploration of supervised pertaining with surrogate labels in material property prediction, advancing methodology and application in the field.

\end{abstract}
\begin{document}

\flushbottom
\maketitle
\thispagestyle{empty}

\section*{Introduction}
%Material discovery is crucial for scientific innovation and a wide range of applications, making it essential to understand the properties of materials

Advancing material discovery is fundamental to driving scientific innovation and expanding application horizons across numerous fields where the need for novel materials becomes incumbent. Such efforts require a deep understanding of material properties to unlock new functionalities and enhance performance in targeted applications~\cite{shen2022high,shahzad2024accelerating,wei2019machine,liu2017materials}.
Traditionally, the scientific community has relied on exhaustive searches through known databases to identify suitable materials, which is costly, time-consuming, prompt to errors, and resource-intensive. 
Introducing first-principles methods, particularly Density Functional Theory (DFT), established a rigorous mathematical framework for predicting material properties~\cite{kohn1965self,hohenberg1964inhomogeneous}. However, this approach remains computationally intensive and time-consuming~\cite{fu2024physics}, and its accuracy is often limited by its approximations, which may only be suitable for some systems.
In recent years, machine learning (ML) and deep learning (DL) models have gained traction in material property prediction, offering faster and more efficient ways to estimate material properties~\cite{bartel2020critical,gong2019predicting,gong2021screening,gupta2021cross,jha2019enhancing,chen2021learning}, at least within the scope of DFT. These models have demonstrated their potential to revolutionize the field by reducing the need for time-consuming calculations.
However, the challenge is that the ML/DL methods heavily depend on large, well-annotated datasets with labels for each material property category. Generating such ground-truth labels often relies on DFT-calculated data, thereby inheriting the same time and cost bottlenecks as traditional methods.

Self-supervised learning (SSL)~\cite{chen2020simple, zbontar2021barlow}, a new learning paradigm in deep learning, provides a promising alternative utilizing unlabeled data and learning useful representations without the need for explicit labels. These models can be pre-trained on vast amounts of unlabeled data to form foundation models, which can then be fine-tuned with smaller labeled datasets. 
These models have been popularly adapted in computer vision and natural language processing, and several algorithms have been proposed, varying according to learning objectives, including BERT~\cite{devlin2019bert}, SimCLR~\cite{chen2020simple}, and BYOL~\cite{grill2020bootstrap}.
This training paradigm has shown its utility in creating models like \textit{ChatGPT} and \textit{DALL·E}, which have gained widespread societal use.
The success of these models reflects SSL's potential in specialized academic research settings and real-world applications that impact society at large.
This approach could have particular relevance within material science as vast databases of material crystals already contain many chemical structures without associated property labels~\cite{jain2013commentary, choudhary2020joint, kirklin2015open}.
Using these untapped data, SSL could unlock significant potential for the prediction of material properties. 

Although self-supervised learning (SSL) has been predominantly explored in vision and language tasks, there have been a few attempts to extend its training paradigm to other scientific disciplines, including molecular chemistry~\cite{zhu2023dual,wang2022improving,zhang2021motif,wang2024evaluating, xu2024triple}.
SSL has shown potential in learning better representations for complex molecular structures by leveraging unlabeled data in these fields. 
However, while molecules and materials share similarities, such as atomic structures and bonding patterns, materials often present more complex, periodic structures.
Unlike molecules, which can be represented by molecular graphs, SMILES strings, or 3D coordinates with well-defined atomic bonding, materials can exhibit crystalline, amorphous, or composite phases that require more intricate representation methods.
For instance, the periodicity of material structures, such as crystal lattices stored in CIF files or their spatial distribution in 3D voxel grids, demands specialized techniques to capture long-range order and periodic boundary conditions. However, molecular structures typically involve covalent bonds between a limited number of atoms and do not exhibit the same degree of complexity in periodicity. As a result, while SSL approaches, such as graph neural networks (GNNs), have proven effective for molecular graphs, adapting these methods to material science requires additional considerations. Techniques incorporating periodic boundary conditions or voxel-based representations are essential to accurately represent material crystals. 
Thus, while there is growing interest in applying SSL to material property prediction, the distinct challenges that material structures pose make it more complex than molecules.

In this study, we advance self-supervised learning (SSL) for material property prediction. Although there have been some early attempts to apply existing SSL setups for this purpose, we take concrete steps forward by proposing \textbf{SPMat (Supervised Pretraining for Material Property Prediction)}, a novel SSL framework that integrates supervisory signals, referred to as surrogate labels, to guide the learning process. Unlike specific labels for each property category, our framework leverages general material attributes (e.g., metal vs. nonmetal) to guide the SSL learning process. To incorporate these supervisory signals, we introduce a novel technique within the standard SSL framework, applicable to both contrastive~\cite{chen2020simple} and noncontrastive~\cite{zbontar2021barlow} approaches. For feature representation learning, we account for the unique characteristics of materials by employing graph neural network (GNN)-based models, specifically the Crystal Graph Convolutional Neural Network (CGCNN)~\cite{cheng2021geometric}. CGCNN effectively encodes local and global chemical information, capturing essential material features such as atomic electron affinity, group number, neighbor distances, orbital interactions, bond angles, and aggregated local chemical and physical properties. Additionally, since SSL frameworks heavily rely on augmentation strategies to enhance learning, we explore existing augmentations and propose
Graph-level Neighbor Distance Noising (GNDN), a novel augmentation approach
that introduces random noise in neighbor distances, further improving model performance. In general, the proposed framework enables us to develop a \textit{foundation model} that demonstrates improved generalization across several material property prediction tasks, outperforming both standard deep learning and existing SSL frameworks.

\begin{figure*}[!t]
\centering
\label{subfig:156}\includegraphics[width=0.95\textwidth]{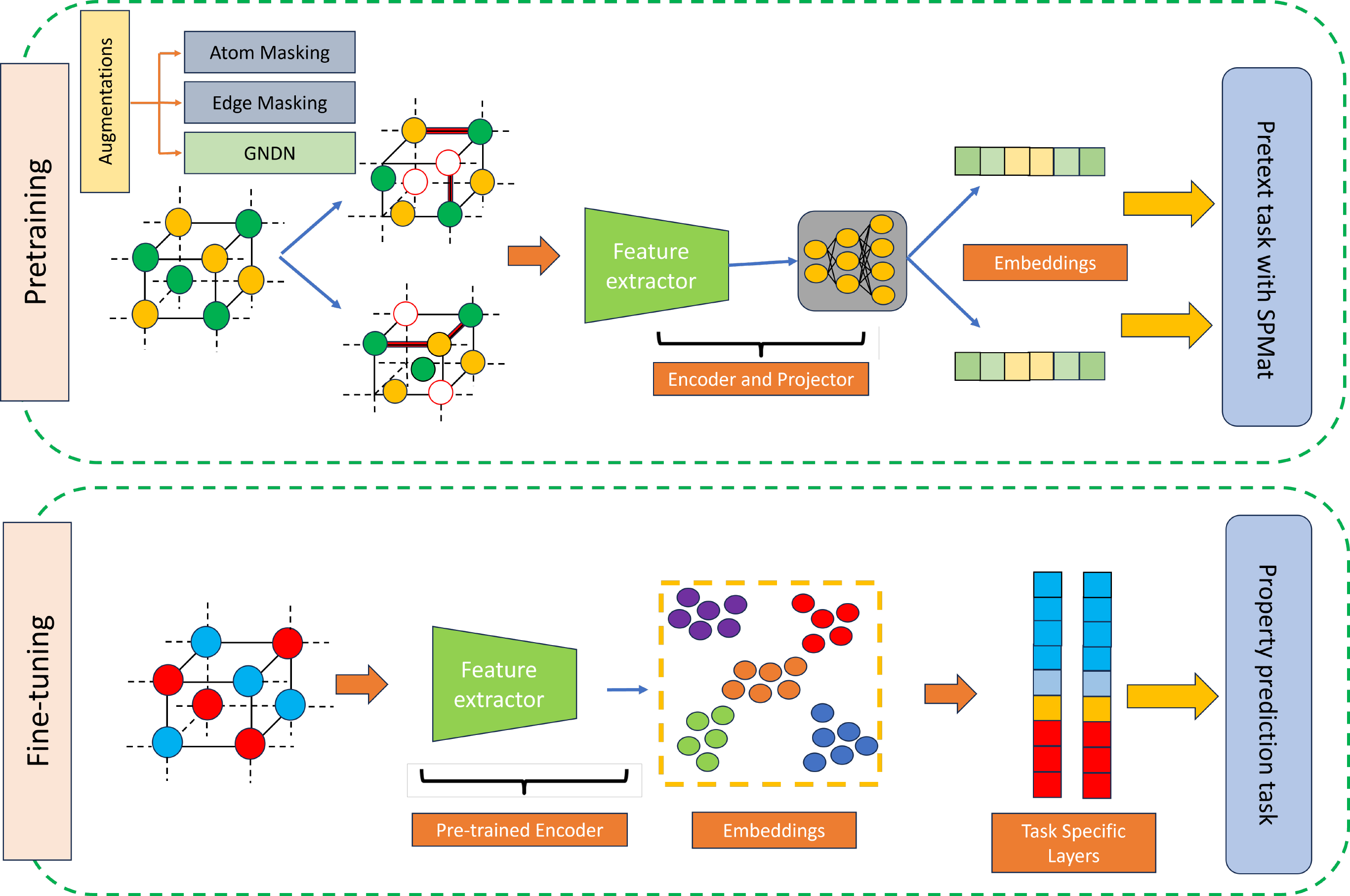}\hfill
\caption{Schematic diagram of the proposed SPMat (Supervised Pretraining for Material Property Prediction) framework. Initially, the material structure undergoes augmentation, including a novel Graph-level Neighbor Distance Noising (GNDN) augmentation, to create diverse views. Deep learning-based encoders and projectors are then used to capture the representations as embeddings, which are utilized in various pretext tasks using labels (top panel). In the downstream task the fine-tuned model starts training on top of the pre-trained encoder and predicts material properties which is a regression task (bottom panel). 
% Block diagram representation of proposed weakly supervised SSL setups for the proposed SP-Mats architecture. For supervised pretraining, the material structure is created from crystallographic information files (CIF) and sequentially augmented to create different views of the same data points. Then the encoders and projectors are used to prepare the embeddings for SSL-based loss function. For Sim-CLR, the embeddings from the same class labels are pulled closer and for different classes pushed apart. On the contrary for the contrastive Barlow-Twin the cross-correlation is increased if the embeddings are from the same class data points and opposite otherwise. Therefore in Figure~\ref{main} in spite of getting an identity matrix we may get a matrix that has off-diagonal elements of same class data points as 1.
}
\label{main}
\end{figure*}
%%%%%%%%%%%%%%%%%%%%%%5
% \begin{figure*}[!ht]
% \centering
% \label{subfig:1567}\includegraphics[width=0.90\textwidth]{fig/main_merge2.pdf}\hfill
% %
% %
% \caption{The original structure created from CIF files goes through sequential augmentation to create different views of the same data points. The augmentation technique used in this research is shown using a block diagram in Fig.~\ref{main1}a. The SSL network is trained with available class labels termed weak labels. The weight and biases of the pre-trained model are shared with fine-tuned models when trained for the property prediction task (Fig.~\ref{main1}b).}
% \label{main1}
% \end{figure*}

% domain the  Moreover, a huge amount of labeled data always assists in the better functioning of ML models in terms of robust predictions. So,
% \begin{itemize}
%     \item Importance of predicting materials
%     \item The usuage of deep learnign and AI for predicting materials and that leading towards pretraining with SSL framework
%     \item Our contribution of doing supervised pretraining
%     \item 
% \end{itemize}

% The Introduction section, of referenced text\cite{Figueredo:2009dg} expands on the background of the work (some overlap with the Abstract is acceptable). The introduction should not include subheadings.

\section*{Results}

\subsection*{Overview of the proposed workflow with SPMat}
% \begin{itemize}
%     \item 
% \end{itemize}
We present the overall schematic of the SPMat framework in Fig.~\ref{main}. SPMat aims to learn material representations using self-supervised learning strategies with supervision with surrogate labels.
% Initially, crystallographic information files (CIFs) are processed to extract the materials' structural information. Weak labels are assigned to individual materials, and the structure is recreated with random augmentations, such as random atom masking and edge masking. 
% \textcolor{green}{
% Subsequently, a graph network is constructed by connecting neighboring atoms based on a distance cutoff and a maximum neighbor count.}
% % \aldo{how atoms are associated? by a distance cutoff?} \abid{Yes Professor, a distance cut-off and a max-neighbor number cut-off}. 
% Random and uniform noise is applied to the neighbor distances to enhance the augmented views further \aldo{no sure what you mean by distance noise = augmented view}. All three augmentations are applied sequentially to generate two distinct instances of the original material structure, facilitating the learning of better representations through diverse versions of the same data point.
% \aldo{maybe this is what you are trying to say: 
Initially, crystallographic information files (CIFs) are processed to extract the structural information of each material. surrogate labels are then assigned to individual materials, and their structures are recreated with random augmentations, including atom masking and edge masking. A graph network is constructed, associating neighboring atoms on the basis of a distance cutoff. To further diversify the augmented representations, we added a novel random, uniform noise-based augmentation to the distances between neighboring atoms. These three augmentations, atom masking, edge masking, and GNDN, are applied sequentially, resulting in two distinct augmented versions of each original data sample. This approach enhances representation learning by exposing the model to varied views of the same material, thereby improving its ability to capture diverse structural features.
The embeddings are generated using a GCNN-based architecture and projectors for contrastive, supervised pretext tasks with surrogate labels. 
We propose two distinct loss objectives: within a minibatch of materials, embeddings from the same data points and those from the same class with randomly augmented views should either be pulled closer (Option 1) or have their correlation maximized (Option 2). In contrast, embeddings from different materials and classes should be pushed apart (Option 1) or made as dissimilar as possible (Option 2). %\aldo{I guess the random distances can be translated to small crystal distortions, what are the values for that? are those distortions for the atomic positions or it is also for the lattice}

To generate diverse versions of data points, we employ a combination of three augmentation techniques: atom masking, edge masking, and a novel graph-level neighbor distance noising (GNDN). Existing methods typically apply spatial perturbations to atomic positions within the original structure of the material by shifting the locations of the atoms to create noisy versions of the material~\cite{magar2022crystal}. However, such perturbations directly alter the crystal structure, potentially affecting key structural properties and undermining the primary objective of augmentation. To address this limitation, we propose graph-level neighbor distance noising (GNDN),
a new
augmentation strategy that introduces noise at the graph level, avoiding direct modifications to the atomic structure. This is achieved by perturbing the distances between neighboring atoms relative to anchor atoms, ensuring the material's core structure remains intact. 
%Formally, the material structure is represented as a graph \( G = (V, E) \), where \( V = \{v_1, v_2, \dots, v_n\} \) denotes the set of nodes (atoms), and \( E \) represents the edges (connections between neighboring atoms). Each edge \( e_{ij} \in E \) is associated with a distance \( d_{ij} = \| \mathbf{r}_i - \mathbf{r}_j \| \), where \( \mathbf{r}_i \) and \( \mathbf{r}_j \) are the position vectors of atoms \( v_i \) and \( v_j \). The noise at the graph level is introduced by modifying \( d_{ij} \) as follows:
%\( d_{ij}' = d_{ij} + \epsilon_{ij} \), where \( \epsilon_{ij} \sim \mathcal{U}(-\delta, \delta) \) represents random noise uniformly distributed within the range \([- \delta, \delta]\), with \( \delta \) controlling the magnitude of the noise. 
This approach preserves the structural integrity of the material while achieving effective augmentation, ensuring the retention of critical properties for downstream tasks (discussed in METHOD section).

% We first describe our proposed augmentation here. Generally, the previous literature~\cite {magar2022crystal} utilized spatial perturbation to atomic positions in the material's original structure by shifting the location of atoms to create a noisy version of the material. However, these perturbations directly modify the crystal structure and can lead to changes in key structural properties hampering the main objective of augmentation. In contrast, in our proposed noising technique, we avoid direct perturbation of the atomic structure by applying noise at the graph level. Therefore, we simulate shifts in neighboring atoms relative to anchor atoms, without altering the material's core structure. 
% This ensures that the material's properties remain intact while achieving the goal of efficient augmentation. 
% We claim that injecting random noise in this way and shifting the neighboring atoms from the anchor atoms is not hampering the material structure but the goal of introducing an efficient augmentation is full-filled. 

The proposed augmentation is utilized within the SPMat training framework, starting with pre-training data \( \mathcal{D} = \left\{ \mathbf{x}_l, \mathbf{y}_l \right\}_{l=1}^{N} \), where \( \mathbf{x}_l \) represents a material crystal, and \( \mathbf{y}_l \) denotes the surrogate label (e.g., magnet vs. non-magnet, metal vs. non-metal). Depending on the type of augmentation, atom masking and edge masking are applied directly to the crystal structure, while the proposed graph-level neighbor distance noising is performed after the crystal is converted into a graph representation. The SPMat framework employs an encoder and a projector to derive \( \mathbf{z}_l \), an embedding representation of the material and its augmented versions.
% The diverse material structures are then subjected to the pretraining where 
% % Formally, 
% we can consider pretraining data $\mathcal{D} = \left\{ \mathbf{x}_l, \mathbf{y}_l \right\}_{l=1}^{N}$ where, $\mathbf{x}_l$ is a material crystal and $\mathbf{y}_l$ is the weak label (e.g., magnet vs. non-magnet, metal vs. non-metal). 
% The SPMat framework employs an encoder and projector to obtain $\mathbf{z}_l$, an embedding representation of the material. 
For any three materials, $\mathbf{x}_i$, $\mathbf{x}_j$ and $\mathbf{x}_k$ and their corresponding surrogate labels $\mathbf{y}_i$, $\mathbf{y}_j$ and $\mathbf{y}_k$, our objective function can be represented as: 
\begin{equation}
\mathcal{L}^{\text{SC}} = \sum_{\substack{\mathbf{z}_{i:1,2},\mathbf{z}_{j:1,2} \\ y_i = {y}_j}} \mathcal{L}^{\text{Attract}}(\mathbf{z}_{i:1,2}, \mathbf{z}_{j:1,2}) + \alpha \sum_{\substack{\mathbf{z}_{i:1,2}, \mathbf{z}_{j:1,2}, \mathbf{z}_{k:1,2} \\ y_i \neq y_k \\ y_j \neq y_k}} (\mathcal{L}^{\text{Repel}}(\mathbf{z}_{i:1,2},\mathbf{z}_{k:1,2}) + \mathcal{L}^{\text{Repel}}(\mathbf{z}_{j:1,2},\mathbf{z}_{k:1,2})) 
\label{simple-sc}
\end{equation}
and, 
\begin{equation}
\mathcal{L}^{\text{BT}} = \sum_{\substack{\mathbf{z}_{i:1,2},\mathbf{z}_{j:1,2} \\ y_i = {y}_j}} \mathcal{L}^{\text{Corr.}}(\mathbf{z}_{i:1,2}, \mathbf{z}_{j:1,2}) + \alpha \sum_{\substack{\mathbf{z}_{i:1,2}, \mathbf{z}_{j:1,2}, \mathbf{z}_{k:1,2} \\ y_i \neq y_k \\ y_j \neq y_k}} (\mathcal{L}^{\text{Decorr.}}(\mathbf{z}_{i:1,2},\mathbf{z}_{k:1,2}) + \mathcal{L}^{\text{Decorr.}}(\mathbf{z}_{j:1,2},\mathbf{z}_{k:1,2})) 
\label{simple-BT}
\end{equation}

Here, $\mathbf{z}_{i:1,2}$, $\mathbf{z}_{j:1,2}$, and $\mathbf{z}_{k:1,2}$ are multiple embeddings corresponding to the augmented views of input materials. The first option, in \ref{simple-sc} (inspired by SimCLR \cite{chen2020simple}) pulls embeddings from the same class surrogate labels closer in the embedding space with $\mathcal{L}^{\text{Attract}}$, while $\mathcal{L}^{\text{Repel}}$ induces repulsion among embeddings of different classes, with $\alpha$ controlling the balance between attraction and repulsion.
For the second option, in \ref{simple-BT} (inspired by Barlow-Twin \cite{zbontar2021barlow}), the correlation of exact class embeddings increases, while the embeddings from different classes are decorrelated. 

To the best of our knowledge, this is the first attempt to add supervision in the pretraining phase of material property prediction. The proposed objective functions are novel and general and have implications for other scientific domains.

% novel contrastive and supervised Barlow-Twin that we propose is also the first of its kind and nowhere to be proposed before this work. We anticipate that providing supervision using class information increases the capacity of the model to learn better representation. Fig.~\ref{main1}(b) depicts the whole architecture of pre-training and fine-tune. The pre-trained model shares its parameters with the downstream regression model that uses 33990 stable material structure and six regression tasks for fine-tuning.  

\subsection*{Application of SPMat to the Materials Project database}
% \begin{itemize}
%     \item discussion on the dataset
%     \item discussion on the weak labels
%     \item discussion on the property predictions
%     \item discussion on diversity (1 or 2 figures inspired by Frontiers)
% \end{itemize}
\begin{figure*}[!ht]
\centering
\label{subfig:669}\includegraphics[width=0.99\textwidth]{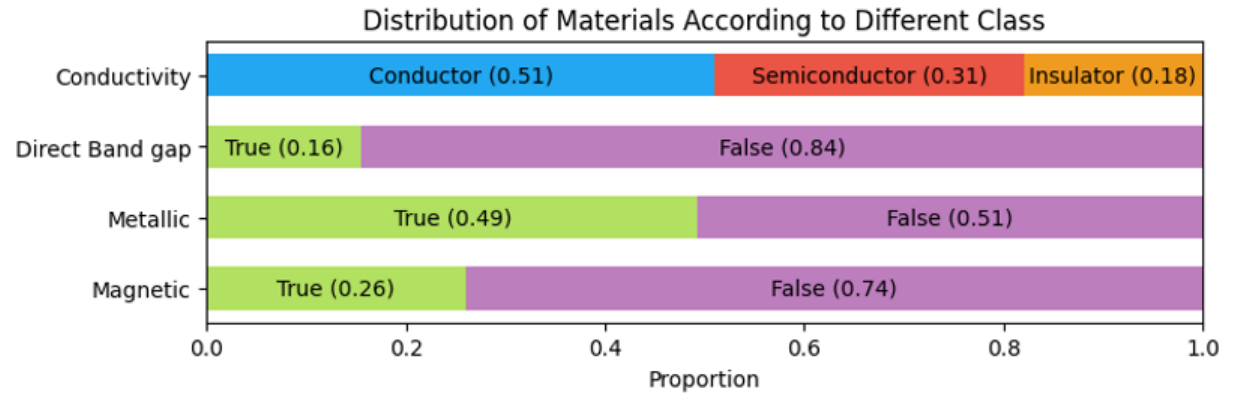}\hfill
\caption{Class-wise distribution of 33,990 stable materials (DFT calculated thermal stability) used in our study %\aldo{how do you define stability here?} 
%\aldo{Again... stable in the sense that they are on the convex? stable because are they thermally stable?... either way, we just need to clarify what we mean by stable, which I guess are materials on the convex hull, right?} \abid{Professor, you are right. I have utilized the stable key to access only the stable materials from MP. This flag is 1 if the material is stable. So far my knowledge goes MP uses
%both the Formation energy and convex hull calculation to determine stability.}
, which shows the proportional breakdown across the main attributes of the materials: conductivity, direct bandgap, metallic, and magnetic. The materials were taken from the Materials Project database~\cite{jain2013commentary}. The following attributes of conductivity are characterized as either true or untrue. Conductivity is divided into three categories: conductor, semiconductor, and insulator. %\aldo{what values were used to define them as semiconductor and insulator?}\abid{Professor, I have taken the standard values, less than .1ev for conductor, in between .1 and 3eV for semiconductor and above that insulator} 
}
     \label{data_distr1}
\end{figure*}
We use the Material Project (MP)~\cite{jain2013commentary} database to curate and preprocess the data set to train and test the proposed supervised pre-training scheme. %\aldo{you sed the database to curate the dataset?} \abid{I had to do some pre-processing like assigning them with insulator, conductor, and semiconductor tags, that's why I mentioned curation}. 
The MP database is a multi-institutional initiative that collects, computes, and stores material and molecular data using density functional theory, mainly using the exchange-correlation functional such as PBE for calculating structure and properties, and hosts information on approximately 155,000 3D material crystals, including significant number of inorganic compounds. %\aldo{still need to clarify how is the dara, meaning, structure PBE, bandgaps R2SCAN?}. 
Notably, bandgaps are often underestimated using PBE, a generalized gradient approximation (GGA) functional that is computationally efficient. To overcome this constraint, MP adds R2SCAN, a meta-GGA functional, to PBE computations to produce more precise predictions of electronic properties such as bandgaps. 
The chemical diversity of the MP database ensures representative samples of various possible arrangements of elements in materials, extending the applicability of our models and results in general. Furthermore, the MP database has been successfully used in numerous previous predictive models to predict different material properties, enhancing its credibility. Therefore, we curated material structures and their corresponding properties from MP to train and verify our models. The properties stored for each material crystal in the MP database are primarily calculated using DFT. As a result, the models trained on these DFT-calculated values aim to predict the material properties as closely as possible to those determined by DFT.

% MP is popular for being authentic and regularly maintained therefore, a lot of studies have used datasets curated from MP to train and test their models. 

In this study, we used 121,000 unstable materials for pre-training and 33,990 stable materials for fine-tuning and validating the proposed framework. 
Our approach introduces a novel element by employing self-supervised pretraining of a GNN using curated class information from the pre-training dataset, which we term \textit{surrogate} labels. These labels include bandgap characteristics as conductor, semiconductor, and insulator, magnetic characteristics as magnet-nonmagnet, metallic characteristics as metal-nonmetal, and bandgap type as direct-indirect gap. %\textcolor{red}{explicitly name these labels}, 
The \textit{surrogate  labels} provide indirect guidance rather than directly influencing the downstream regression task. Consequently, we refer to this process as \textit{supervised pertaining with surrogate labels}.
To evaluate our approach, we predicted six fundamental material properties: Formation energy, electron bandgap, material density, Fermi energy, energy per atom, and atomic density, as obtained from the DFT calculations in the MP database. These properties are critical for understanding a material’s potential in scientific discovery, academic research, and industrial applications. 
While pretraining leverages unrelated class labels, our ultimate goal is to predict distinct material properties with significant practical implications. This distinction underlines our use of supervision in the pre-training phase, where the labels are not related to the downstream prediction tasks.
The chemical diversity of our pretraining dataset is shown in Fig.~\ref{data_distr1}, comprising 51\% conductors, 31\% semiconductors, and 18\% insulators, with an equal mixture of metallic and non-metallic materials and a higher proportion of magnetic materials, highlighting the diversity of the data set.

\begin{figure*}[!t]
\centering
\label{subfig:670}\includegraphics[scale=0.5]{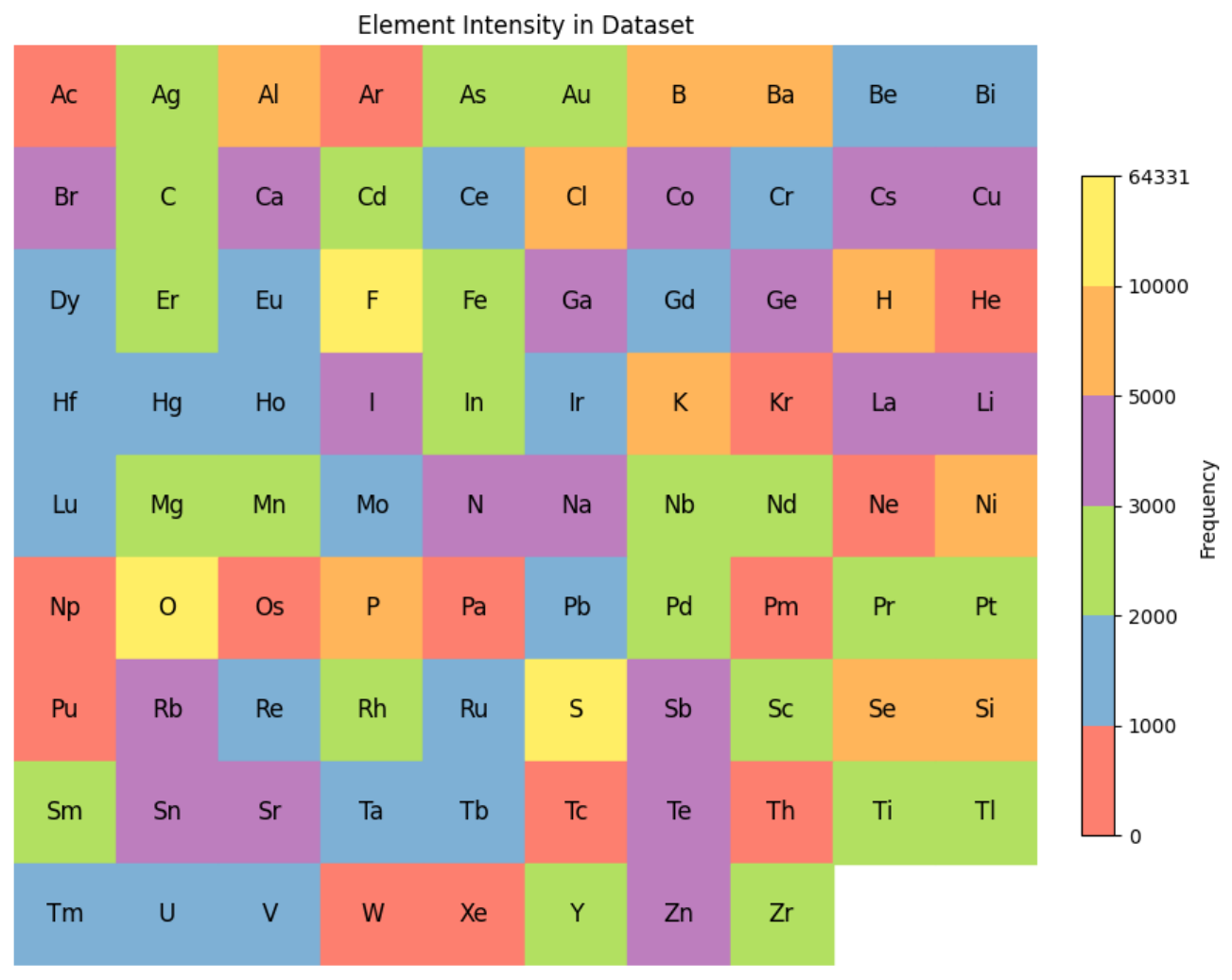}\hfill
\caption{Distribution of element frequencies in the dataset used in our tests. The most frequent element is oxygen (O), whereas the least common element is neon (Ne). The color scale shows the element frequency, while considerable diversity is indicated by a Shannon entropy of 3.81.}
     \label{data_distr2}
\end{figure*}

We also looked at the elementwise frequency to analyze the chemical diversity of the fine-tuning dataset. It comprises 88 unique elements, visualized in Fig.~\ref{data_distr2}, where tile colors indicate frequency (red = lowest, yellow = highest). Oxygen (O) is the most common, appearing 64,331 times, reflecting its frequent presence in inorganic compounds, while neon (Ne) is the least common, appearing once, consistent with the chemical inertness of noble gases. The data set shows moderate diversity, with a Shannon entropy of 3.81, indicating that while various elements are present, some, such as oxygen, dominate.

\subsection*{SPMat generates better representation}
\begin{figure*}[!ht]
\centering
\label{subfig:555}\includegraphics[width=0.98\textwidth]{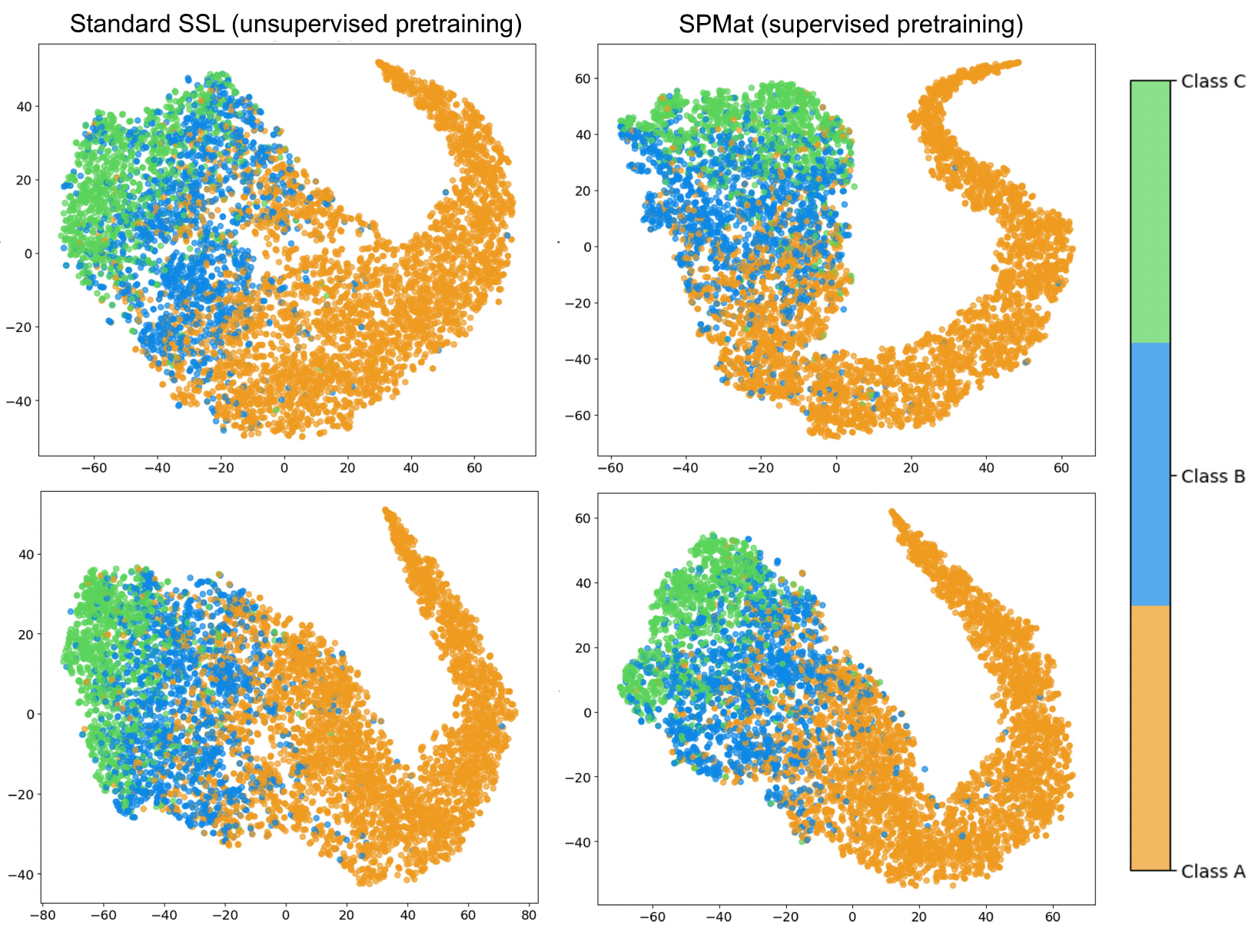}\hfill
\caption{t-SNE visualization of embeddings from SPMat compared to standard SSL models (Barlow Twins and SimCLR) for bandgap prediction.
The left panel shows embeddings from the unsupervised SSL models: Barlow Twins (top) and SimCLR (bottom). The right panel shows SPMat's results for the two variants defined in Eqn.\ref{simple-BT} (top) and Eqn.\ref{simple-sc} (bottom). Colors represent the three classes obtained by discretizing the bandgap values.
}
\label{tsne}
\end{figure*}
%%%%%%%%%%%%%%%%%%%%%%%%%%%%%%%%%%%%%%%%%%%%%
\begin{figure*}[!ht]
\centering
\label{subfig:575}\includegraphics[width=0.98\textwidth]{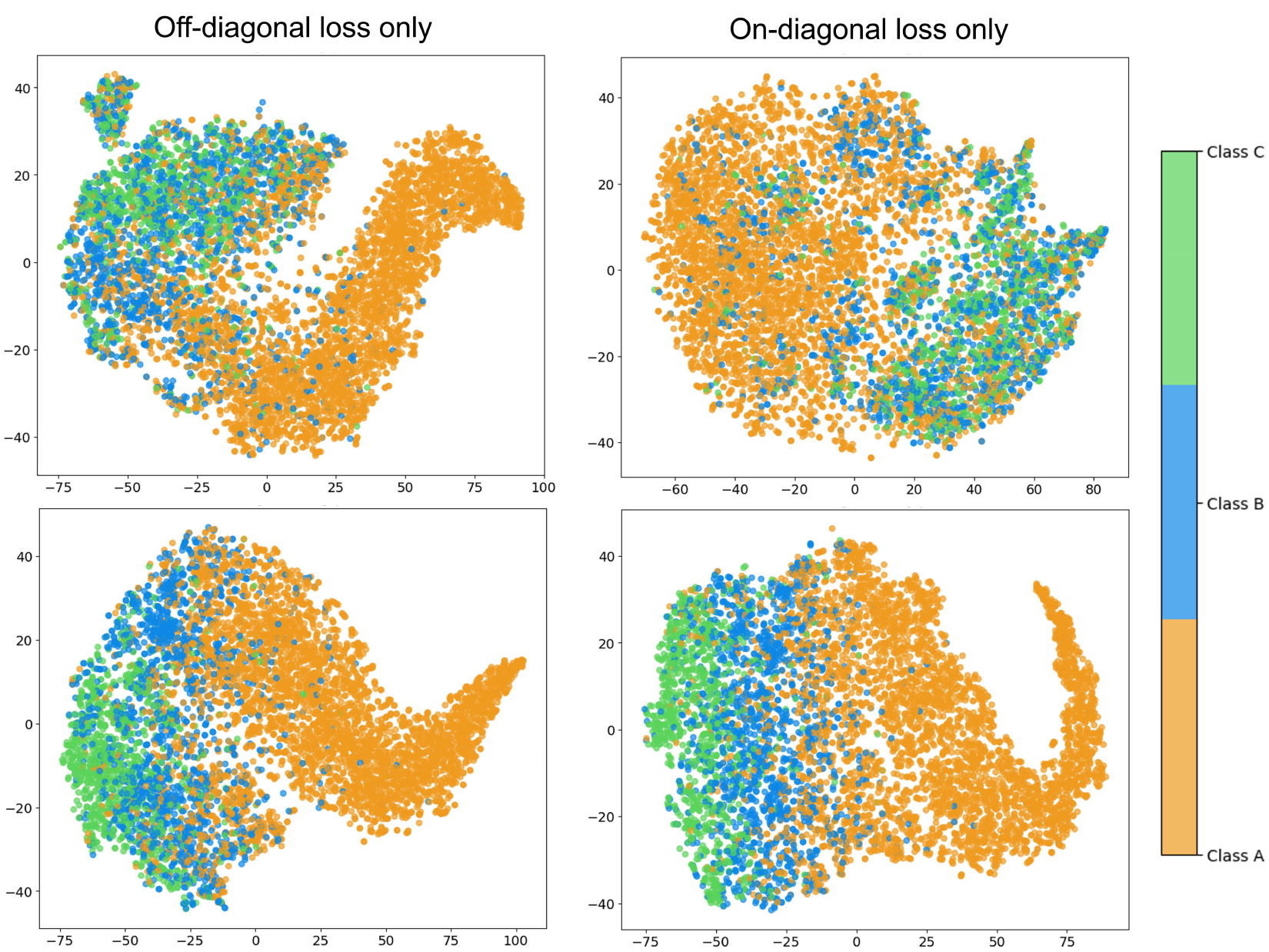}\hfill
\caption{t-SNE visualization for the supervised Barlow Twins loss, keeping only the on- and off-diagonal portions of the loss for pre-trained and fine-tuned models. The left panel represents the model trained with only the off-diagonal loss, while the right panel shows the model trained with only the on-diagonal loss. In each case, the t-SNE plots for pre-trained models are shown on the top and for fine-tuned models on the bottom.}
\label{tsne2}
\end{figure*}
%%%%%%%%%%%%%%%%%%%%%%%%%%%%%%%%%%%%%%%
\begin{figure*}[!ht]
\centering
\label{subfig:578}\includegraphics[width=0.99\textwidth]{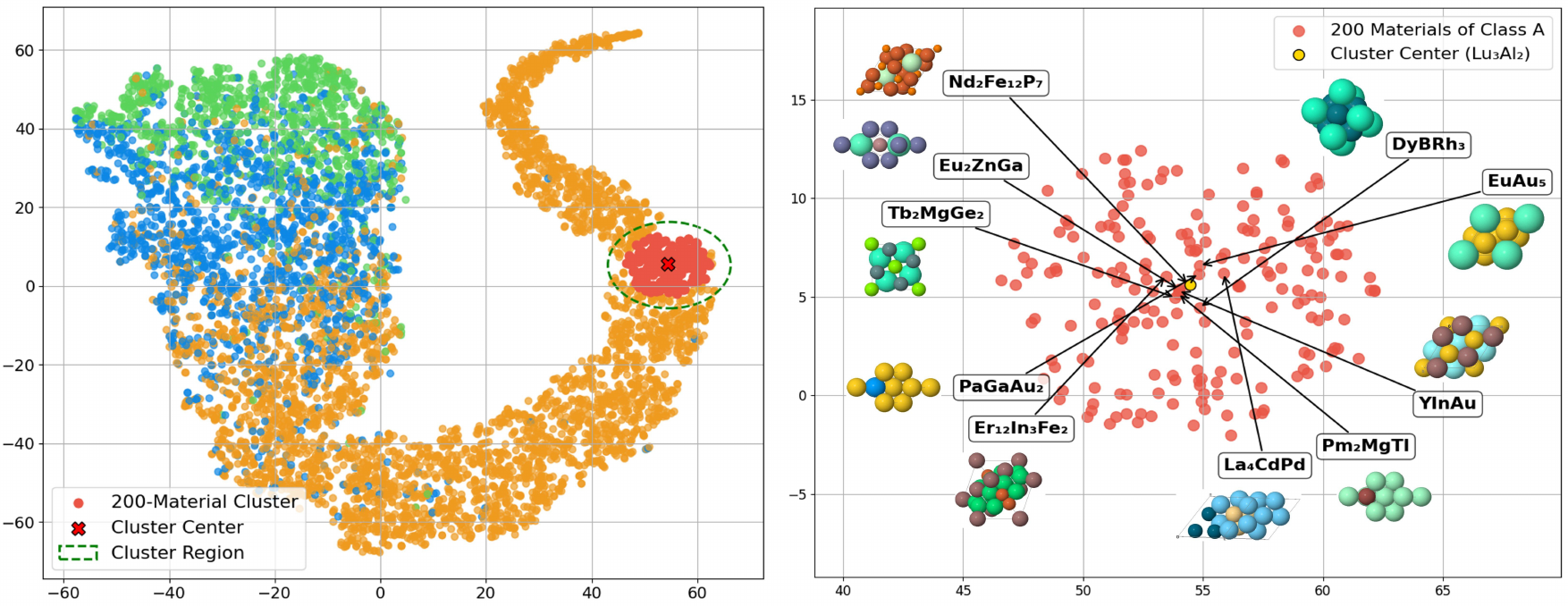}\hfill
\caption{Chemical and structural significance from t-SNE plots, showing how the model captures different attributes. The left panel shows an embedding cluster of 200 materials with the lowest Euclidean distances from the cluster center. The right panel shows ten arbitrarily chosen members of the cluster to analyze their structural and chemical similarities.}
\label{tsne3}
\end{figure*}
%%%%%%%%%%%%%%%%%%%%%%

In Fig.\ref{tsne}, we present t-SNE plots\cite{van2008visualizing} obtained from the embedding representations of the fine-tuned SPMat model compared to the standard fine-tuned SSL models. In this specific case, the models are fine-tuned to predict bandgap values. In the figure, the left panel represents the baselines (i.e., standard SSL models), and the right panel represents SPMat. For both SPMat and the baseline, we show two types of SSL training: Barlow Twins for the baseline and Eqn.\ref{simple-BT} on top and SimCLR for the baseline and Eqn.\ref{simple-sc} on the bottom.
For enhanced visualization, we group the predicted bandgap values into three classes: Class A (bandgap $\in [0, 1]$), Class B (bandgap $\in (1, 3]$), and Class C (bandgap $\in (3, 5]$). As shown in the figure, the results in the right panel consistently demonstrate better separability. For the baselines, while the clusters corresponding to the classes are visible, there is considerable overlap, particularly between class B (blue) and class A (orange), indicating that the embeddings do not ideally separate the classes in the latent space. The clustering for class C (green) is somewhat more distinct; however, significant mixing between the classes remains, suggesting that the embeddings are not fully optimized for class separability.
This enhanced separability confirms that the proposed SPMat, by leveraging supervision using surrogate labels during pretraining, results in meaningful and discriminative embeddings that better capture the material properties of the data set.

Between the bottom-right panel (SPMat-SC) and the top-right panel (SPMat-BT), the latter generates a better t-SNE plot in terms of separability. The SPMat-SC loss encourages same-class embeddings to be closer via a contrastive framework. However, due to its softmax-like structure, it emphasizes relative similarity rather than enforcing strong absolute similarity within classes. It also does not address redundancy, so embeddings within a class may still be correlated or repetitive. As a result, while same-class samples are grouped, the clusters appear more diffuse, as seen in the Fig. \ref{tsne} bottom-right panel (e.g., overlapping between orange and blue classes).
In contrast, SPMat-BT (Eqn. \ref{simple-BT}) explicitly maximizes similarity across all same-class sample pairs, including augmented views and different samples. This encourages tighter intra-class clustering. Additionally, the second term in Eqn. \ref{simple-BT} reduces redundancy by minimizing inter-class similarity, helping generate more disentangled and compact class clusters (top right panel).
This results in noticeably tighter clusters for each class, with minimal spread within each group (top right plane).

We analyze the impact of the individual components of Eqn. \ref{simple-BT} by pretraining models separately using the on-diagonal loss (which increases within-class correlation) and the off-diagonal loss (which reduces between-class correlation) to assess the effects of each configuration on clustering quality.
Figure~\ref{tsne2} shows the resulting t-SNE plots: the left panel for off-diagonal loss, and the right for on-diagonal loss (top: pretrained, bottom: fine-tuned).
The on-diagonal loss alone leads to significant class overlap and diffuse clusters due to the lack of inter-class separation, making embeddings less useful for bandgap prediction. Fine-tuning improves this somewhat but remains suboptimal. In contrast, the off-diagonal loss achieves better class separation, particularly for Classes B and C, though Class A remains elongated, demonstrating the need for intra-class compactness. These results underscore the role of inter-class dissimilarity in learning useful representations and suggest that combining both loss terms, as done in SPMat (Eqn. \ref{simple-BT}), leads to more compact and discriminative embeddings (Fig.~\ref{tsne}).
% The outcome of the pretrained and fine-tune models for both the setups are visualized in the Fig.~\ref{tsne2}. In the left plane of the Fig.~\ref{tsne2}, the t-SNE embeddings for pretrained and fine-tuned model for off-diagonal loss are shown. In the right plane, embeddings of pretrained and fine-tuned models trained only on on-diagonal loss are demonstrated. The right panel (top: pretrained, bottom: fine-tuned) shows that the on-diagonal loss results in heavy overlap between all three classes, with diffuse clusters and an uninterpretable structure. This is because the on-diagonal loss lacks a mechanism to push different classes apart, leading to entangled embeddings that are non-discriminative for bandgap prediction. However, the fine-tuning helps to get a better class separation compared to the pretrained model. The off-diagonal loss outperforms the on-diagonal loss by achieving clear separation among classes, which is critical for distinguishing between bandgap classes. The compactness of the Class C and Class B clusters in the off-diagonal plots is a byproduct of separation, though the elongated Class A cluster highlights the need for within-class optimization. These results demonstrate the importance of between-class dissimilarity in learning discriminative embeddings and also suggest that a combined loss (balancing both terms) could further improve compactness, as demonstrated in our proposed SPMat Eqn. \ref{simple-BT} (Fig.~\ref{tsne}).
\newline
\textbf{Interpretable representations:} To understand the interpretability of our methodology, we present a visual illustration in Fig.~\ref{tsne3}, offering an in-depth examination of the t-SNE clustering of material embeddings. This figure highlights a cluster of 200 materials with the lowest Euclidean distances from a randomly chosen cluster center, Lu$_3$Al$_2$. It demonstrates how  our approach captures structural and chemical similarities among materials. The model's effectiveness in material representation learning is validated by visualizing and analyzing the nearest materials in Euclidean space, offering insights into its ability to encode meaningful material features.

The left panel displays the 200 closest materials (red points) to the cluster center (yellow star, Lu$_3$Al$_2$), along with the cluster boundary (green outline). The right panel provides a zoomed-in view of this cluster, showing a dense grouping that reflects the embeddings’ ability to capture consistent material properties. The right panel also focuses on the ten materials closest to the cluster center—YInAu, Nd$_2$Fe$_{12}$P$_7$, Eu$_2$ZnGa, PaGaAu$_2$, Pm$_2$MgTl, Tb$_2$MgGe$_2$, EuAu$_5$, Er$_{12}$In$_3$Fe$_2$, DyBRh$_3$, and La$_4$CdPd—each connected by lines to emphasize their proximity in the embedding space.

Beyond proximity in embedding space, we examine the shared chemical and structural attributes of the eleven clustered materials.
This analysis further supports the coherence observed in the t-SNE plots. The materials exhibit overlapping valence and conduction bands, confirming their metallic nature, and share a common compositional motif involving rare earth (Y, Nd, Eu, Pm, Tb, Er, Dy, La) or actinide (Pa) elements, combined with transition metals (Fe, Zn, Au, Rh, Pd) and p-block elements (In, P, Ga, Tl, Ge, Cd, B).
Structurally, they align with known types such as 
Heusler-like (YInAu, La$_4$CdPd), ThMn$_{12}$-type (Nd$_2$Fe$_{12}$P$_7$), Zintl or Zintl-like phases (Eu$_2$ZnGa, Pm$_2$MgTl, Tb$_2$MgGe$_2$), and RhB$_3$-type (DyBRh$_3$), all characterized by high-coordination, close-packed structures that favor metallic bonding. The embeddings effectively capture these similarities, as reflected in the tight clusters. Several materials—e.g., Nd$_2$Fe$_{12}$P$_7$, Er$_{12}$In$_3$Fe$_2$, DyBRh$_3$—also likely exhibit ferromagnetic or paramagnetic behavior due to the presence of magnetic RE elements and Fe. Notably, the ThMn$_{12}$-type structure in Nd$_2$Fe$_{12}$P$_7$ is analogous to that of Nd$_2$Fe$_{14}$B, a known ferromagnet, suggesting that the embeddings may also encode magnetic interactions.
These observations suggest that the improved interpretability of our proposed SSL method—reflected in its class-wise separability—enables deeper insights into the underlying chemical, structural, and magnetic patterns captured in the learned representations.

\subsection*{SPMat achieves improved generalization for property predictions}
% \begin{itemize}
%     \item Figure 1
%     \item Table 1
%     \item Table 2
% \end{itemize}
%
%
\begin{figure*}[!htbp]
\centering
\label{subfig:56}\includegraphics[width=0.98\textwidth]{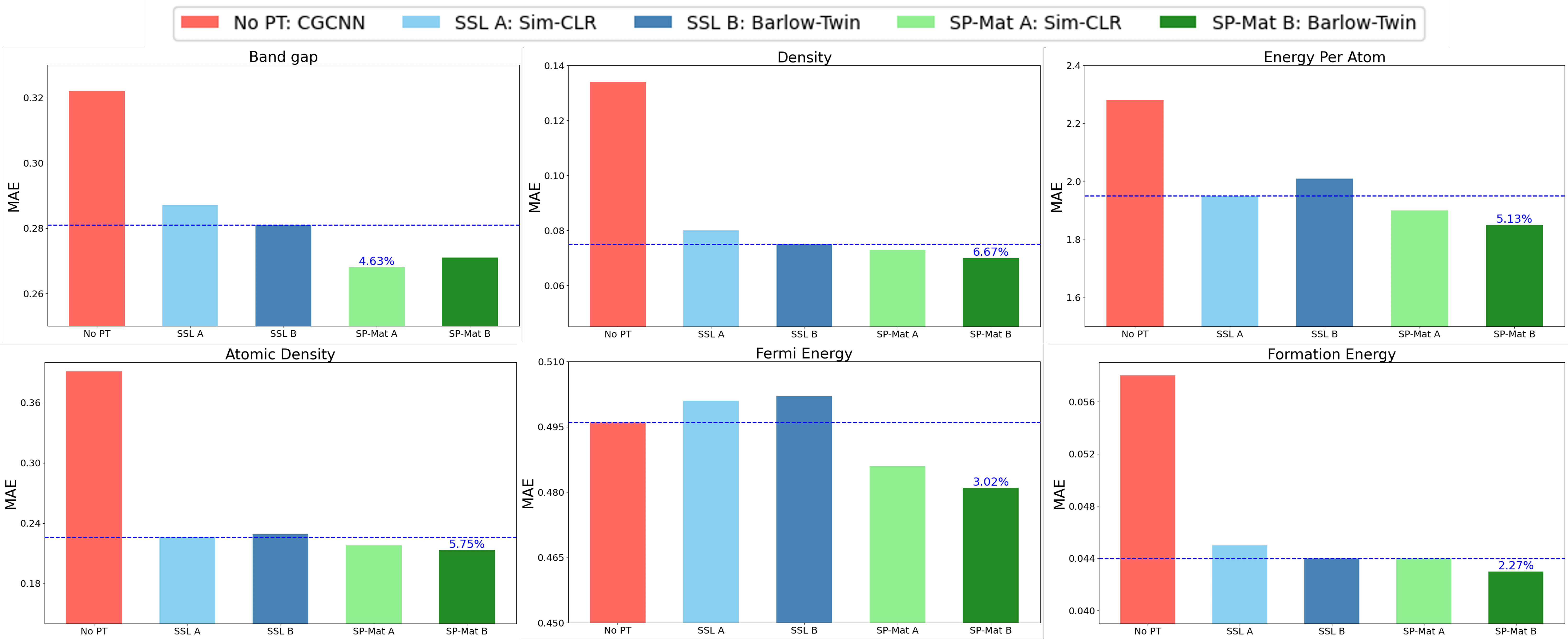}\hfill
\caption{Comparison of proposed supervised SPMat A: SimCLR and SPMat B: Barlow-Twin models with the supervised baseline No PT: CGCNN, SSL A: SimCLR, and SSL B: Barlow-Twin. No PT: CGCNN represents the CGCNN model. SSL A and SSL B demonstrate the baseline SimCLR and Barlow-Twin, while SPMat A and B show the MAE results for fine-tuned models with supervised pertaining. For clarity, blue shades have been given to the bars representing no supervision during pretraining; conversely, green shades are provided to the bars of supervised models. The dashed blue line signifies the lowest MAE value among the No PT, SSL A, and SSL B models, while the percent improvement of our proposed best-performing model is labeled above its bar.
}
%\textcolor{blue}{1. remove "pretraining configurations" 2. reduce the gap between each bar by about 40 percent. }}
\label{model_1}
\end{figure*}
%%%%%%%%%%%%%%%%%%%

In this section, we present the key findings of this study, focusing on evaluating SPMat’s performance in property prediction and comparing it with existing methods.
As discussed earlier, SPMat incorporates surrogate supervision during pretraining. In our study, we utilized four types of bluesurrogate labels: bandgap (BG), metal-non-metal (Met), magnetic-nonmagnetic (Mag), and gap directness (Gap).
These four surrogate supervisory signals are applied to two variants of SPMat, referred to as SPMat-Barlow-Twin (Eqn.\ref{simple-BT}) and SPMat-SimCLR (Eqn.\ref{simple-sc}), resulting in a total of eight different setups for comparison.
The pretrained models from surrogate supervision are fine-tuned to predict six prevalent properties: Formation energy per atom (FE), bandgap (BG), density (density), Fermi energy (EFermi), energy per atom (EPA) and atomic density (A. density).
For comparison, we consider CGCNN (a baseline without pre-training) and two state-of-the-art SSL models: SSL-Barlow-Twin and SSL-SimCLR, representing standard SSL training without supervision as reported in Crystal-Twins~\cite{magar2022crystal}. 
We note that Crystal-Twins is a pioneering SSL framework for crystalline material property prediction and is considered a state-of-the-art (SOTA) model in this domain. Although we refer to Crystal-Twins-based Barlow Twins and SimCLR as baselines in our study, their novelty and strong performance position them among the leading SSL methods for materials science. Our proposed SPMat framework builds on this foundation by introducing supervision into SSL, using supervised contrastive learning and a supervised version of Barlow Twins, both with CGCNN as the backbone encoder. To ensure a fair and meaningful comparison, we benchmark against Crystal-Twins-based SSL models (SimCLR and Barlow Twins) and the supervised CGCNN baseline, all of which share a similar CGCNN backbone, isolating the effect of incorporating supervision.
% \textcolor{blue}{Crystal Twins framework is a pioneering SSL approach for crystalline material property prediction and can be considered a SOTA model in this context. While we refer to Crystal Twins-based Barlow-Twins and Sim-CLR as a baseline in our study, it's novelty and performance position it among the leading SSL methods in the field. Our proposed SPMat framework, which builds on supervised SSL using SupCon and Supervised Contrastive Barlow-Twins with CGCNN as the backbone encoder, is designed to explore the potential of supervision within this SSL paradigm. Consequently, we chose to benchmark against SSL models like Crystal-Twin-based SimCLR and Barlow-Twins, as well as the supervised CGCNN baseline, all of which share a comparable backbone (CGCNN). This ensures a fair and meaningful comparison that isolates the contribution of our supervised SSL approach.}

In Fig.~\ref{model_1}, we present the results for the six properties, comparing the SPMat variants—SPMat-SimCLR (SPMat A) and SPMat-Barlow-Twin (SPMat B)—with existing models. CGCNN (No PT), SSL-SimCLR (SSL A), and SSL-Barlow-Twin (SSL B).
For this analysis, SPMat is pre-trained with the surrogate BG label. As shown, across the six properties, the lowest MAE is consistently achieved by our proposed framework. A blue dotted line is added to indicate the model that performs the best among the existing methods, along with the percentage improvement achieved by SPMat. In particular, SPMat shows improvements ranging from over 2\% to up to 6.67\%, highlighting its robustness in achieving enhanced generalization for these critical properties.

In Table~\ref{tab1} and Table~\ref{tab2}, we present detailed results for all surrogate labels used during the pre-training of SPMat, comparing it with CGCNN and SSL-BT in Table~\ref{tab1}, and against CGCNN and SSL-SimCLR in Table~\ref{tab2}. The best results are highlighted in bold in both tables and the second-best results are underlined. For the baseline SSL models, atom and edge masking are used as augmentation as proposed in the original architecture, while GNDN-based augmentation is applied on SPMat setups. In particular, we discuss the effect of augmentation in Table~\ref{tab:effect_aug} for the baseline and our proposed setup.
As shown in Table~\ref{tab1}, the SSL-BT model outperforms the baseline CGCNN model for all properties except Fermi Energy, demonstrating the advantages of SSL pretraining. With our proposed SPMat-BT (SPMat with Barlow-Twins-based training), the results consistently outperform both the no-pre-training baseline and SSL training across all cases. Among the surrogate labels, the bandgap label achieves the best results in four out of five configurations and ranks second in one configuration.
Similarly, in Table~\ref{tab2}, which presents the results for SimCLR-based models, all SPMat variants achieve superior performance compared to baselines. Among the surrogate labels, the Is-Metal label achieves the best results for three properties, while the bandgap and Is-Magnetic label achieve the best results in two different settings.
Overall, across all six properties and under various surrogate labels, these results demonstrate the ability of our proposed SPMat variants to provide improved generalization.
\begin{table}[!t]
\begin{adjustbox}{max width=\textwidth}
    \begin{minipage}{\textwidth}
    \renewcommand{\arraystretch}{1}
    \setlength{\tabcolsep}{8pt}
        \centering
        \caption{Comparison of MAE values for six properties for self-supervised Barlow-Twin and our proposed novel supervised contrastive Barlow-Twin for surrogate labels Bandgap (BG), Is-Metal(Met), Is-Magnetic (Mag), and Is-Gap-Direct (Gap).}
        \begin{tabular}{cccccccccc}
            \toprule
            \toprule
            \textbf{Property}   & \textbf{Formation Energy} & \textbf{Bandgap} & \textbf{Density} & \textbf{Fermi Energy} & \textbf{Energy Per Atom} & \textbf{A. Density} \\
            \textbf{\# Crystals} & \textbf{33,990} & \textbf{33,990} & \textbf{33,990} & \textbf{33,990} & \textbf{33,990} & \textbf{33,990} \\
            \midrule
            \textbf{CGCNN}   & 0.058 & 0.322 & 0.134 & 0.496 & 2.281 & 0.391 \\
            \textbf{SSL-BT}   & \underline{0.044} & 0.281 & 0.075 & 0.502 & 2.01 & 0.229 \\
            \textbf{SPMat-BT-BG}    & \textbf{0.043} & 0.271 & \textbf{0.070} & \textbf{0.481} & \textbf{1.85} & \underline{0.213} \\
            \textbf{SPMat-BT-Met}& 
           \underline{0.044}& 0.271 & 0.072 & 
            \underline{0.491} & \textbf{1.85} & \textbf{0.212} \\
            \textbf{SPMat-BT-Mag}    & \textbf{0.043} & \textbf{0.266} & \underline{0.071} & \underline{0.491} & \underline{1.87} & \underline{0.213} \\
            \textbf{SPMat-BT-Gap}  & \textbf{0.043} & \underline{0.270} & 0.072 & 0.496 & \underline{1.87} & 0.219 \\
            \bottomrule
            \bottomrule
        \end{tabular}
        \label{tab1}
        \caption*{In this table, the best MAE result for any particular property is bold, while the second best is underlined. Supervision with surrogate label is applied in the broader material class as metal-nonmetal, magnetic-nonmagnetic, insulator-conductor-semiconductor, and gap-direct-not direct.}
    \end{minipage}
\end{adjustbox}
\end{table}
\begin{table}[!t]
\begin{adjustbox}{max width=\textwidth}
    \begin{minipage}{\textwidth}
    \renewcommand{\arraystretch}{1}
    \setlength{\tabcolsep}{8pt}
        \centering
        \captionof{table}{Comparison of MAE values for six crucial properties from MP for self-supervised SimCLR and our proposed supervised contrastive SimCLR for class labels Bandgap (BG), Is-Metal(Met), Is-Magnetic (Mag), and Is-Gap-Direct (Gap).}
        \begin{tabular}{cccccccccc}
            \toprule
            \toprule
            \textbf{Property}   & \textbf{Formation Energy} & \textbf{Bandgap} & \textbf{Density} & \textbf{Fermi Energy} & \textbf{Energy Per Atom} & \textbf{A. Density} \\
            \textbf{\# Crystals} & \textbf{33,990} & \textbf{33,990} & \textbf{33,990} & \textbf{33,990} & \textbf{33,990} & \textbf{33,990} \\
            \midrule
            \textbf{CGCNN}   & 0.058 & 0.322 & 0.134 & 0.496 & 2.28 & 0.391 \\
            \textbf{SimCLR}   & 0.045 & 0.287 & 0.080 & 0.501 & 1.95 & 0.226 \\
            \textbf{SPMat-SC-BG}    & 0.044 & \underline{0.268} & \underline{0.073} & \textbf{0.486} & \underline{1.90} & \textbf{0.218} \\
            \textbf{SPMat-SC-Met}& 
           0.044& \textbf{0.267} & \textbf{0.072} & 
            0.494 & \underline{1.90} & \textbf{0.218} \\
            \textbf{SPMat-SC-Mag}    & \textbf{0.042} & 0.276 & \textbf{0.072} & \underline{0.491} & 1.95 & 0.226 \\
            \textbf{SPMat-SC-Gap}  & \underline{0.043} & 0.279 & 0.074 & 0.492 & \textbf{1.79} & \underline{0.219} \\
            \bottomrule
            \bottomrule
        \end{tabular}
        \label{tab2}
        \caption*{In this table, the best MAE result for any particular property is bold, while the second best is underlined. Supervision with surrogate labels is applied in the broader material class as metal-nonmetal, magnetic-nonmagnetic, insulator-conductor-semiconductor, and gap-direct-not direct.}
    \end{minipage}%
\end{adjustbox}
\end{table}

\begin{table}[!t]
    \centering
    \captionof{table}{The effect of proposed new augmentation on top of Atom and Edge masking on MAE values of the six predicted properties for supervised Barlow-Twin (BT) and SimCLR with bandgap as a surrogate supervisory signal.}
    \renewcommand{\arraystretch}{1.5}
    \setlength{\tabcolsep}{6.5pt}
    \begin{tabularx}{\textwidth}{ccccccccc}
        \toprule
        \multicolumn{1}{l}{SSL} &Method &Augmentation & Formation Energy & Bandgap & Density & E Fermi & EPA & A Density \\ \hline
        \multirow{4}{*}{BT} &\multirow{2}{*}{SSL} &Atom+Edge+GNDN & .044 & .276 & .074 & \underline{.486} & 1.91 & .225 \\ \cline{3-9}&
                                   & Atom+Edge & .044 & .281 & .075 & .502 & 2.01 & .229 \\ \cline{2-9}
                                  
          &\multirow{2}{*}{SPMat} &Atom+Edge+GNDN & \textbf{.043} & \underline{.271} & \textbf{.070} & \textbf{.481} & \textbf{1.85} & \textbf{.213} \\ \cline{3-9}&
                                   & Atom+Edge & .044 & .279 & .077 & .498 & 1.98 & .221 \\ \hline
                                   
        \multirow{4}{*}{SimCLR} &\multirow{2}{*}{SSL} & Atom+Edge+GNDN & .044 & .274 & .074 & .490 & 1.92 & .229 \\ \cline{3-9}&
                                   & Atom+Edge & .045 & .287 & .080 & .501 & 1.95 & .226 \\ \cline{2-9}

         &\multirow{2}{*}{SPMat} & Atom+Edge+GNDN & .044 & \textbf{.268} & \underline{.073} & \underline{.486} & \underline{1.90} & \underline{.218} \\ \cline{3-9}&
                                   & Atom+Edge & .044 & .283 & .074 & .502 & 1.93 & .227 \\ 
                                   \bottomrule
    \end{tabularx}
    \caption*{Comparative results of supervised Barlow-Twin and SimCLR with and without the proposed new augmentation technique for stable materials curated from the MP database. The best result for any particular property has been bolded, and the second best underlined.}
    \label{tab:effect_aug}
\end{table}

\subsection*{Effects of new augmentation}
This article introduces a new random noise-based augmentation for the SSL GNDN training, which is incorporated into the SPMat pipeline. In this section, we evaluate the impact of this new increase on both standard SSL pre-training and the proposed SPMat with supervised pre-training.
The results of this evaluation are presented in Table~\ref{tab:effect_aug}, comparing both variants of SPMat with the baseline SSL models (Barlow-Twin and SimCLR). Since some form of augmentation is always necessary, we use Atom Masking and Edge Masking as the baseline augmentations and assess the effect of adding the proposed GNDN augmentation.  GNDN augmentation involves applying random and uniform noise to neighbor distances.

First, for standard SSL models, the addition of our novel augmentation improves results across 11 of 12 settings. The only exception is the Formation Energy property, where SSL-BT achieves the same result even after adding the proposed augmentation.
In the case of SPMat, the trend is consistent: in every instance, the inclusion of GNDN augmentation significantly reduces the MAE values compared to using Atom Masking and Edge Masking alone.
Overall, these results demonstrate that incorporating random noise as an augmentation improves prediction accuracy in both the SPMat and baseline setups, surpassing the performance achieved using only Atom and Edge Masking.

\subsection*{Effect of batch size on SPMat}
% Here, we discuss the effect of changing batch-size during pretraining our supervised Barlow-Twin setup. 
% To test the effect of batch size on downstream tasks like material property prediction, we use bandgap as a supervising class label. In establishing the original results we have used 128 as the batch size for the Barlow-Twin setup. Therefore, this time, we compare our results for the batch sizes 64, 128, and 256. The results for testing the effect of changing batch size in the supervised Barlow-Twin setup are given in the Table\ref{tab:batchsize_effect}. It is seen that the effect of increasing batch size is not significant for supervised Barlow-Twin as it is for contrastive losses like Sim-CLR. The performance of the proposed supervised Barlow-Twin slightly increases or remain the same with the increase in batch size.  Since our loss is supervised, it relies on having both same-class (positive) and different-class (negative) pairs in the batch. Therefore, larger batches increase the likelihood of including multiple classes and multiple samples per class, enhancing discrimination and improving performance by providing more diverse class interactions, a greater number of pairwise terms, and more stable optimization. However, the effect of changing batch size is not severe for supervised Barlow-Twin in comparison to SimCLR for its matrix-based design, which makes it more flexible, maintaining reasonable performance even with smaller batches.
To test the effect of batch size on downstream tasks like material property prediction, we use bandgap as the supervising class label. In the original setup, we used a batch size of 128 for the SPMat.
% the supervised Barlow Twins model. 
Here, we compare results across batch sizes of 64, 128, and 256 for the SPMat-BT. Table~\ref{tab:batchsize_effect} shows the results. We observe that increasing batch size has a limited effect on supervised Barlow Twins, unlike contrastive losses such as SimCLR. The performance either slightly improves or remains stable with larger batches.
Since our loss is supervised, larger batches increase the likelihood of including multiple classes and samples per class, improving discrimination through more diverse pairwise terms and more stable optimization. However, due to its matrix-based design, supervised Barlow Twins remains more flexible and maintains strong performance even with smaller batches, unlike SimCLR.

%%%%%%%%%%%%%%%%%%%%%%%%%%%%%%%%%%%%%%%
\begin{table}[!t]
    \centering
    
    \captionof{table}{The effect of varying batch size on MAE values of the six predicted material properties for supervised Barlow-Twin configuration.}
    \renewcommand{\arraystretch}{1.5}
    \setlength{\tabcolsep}{9pt}
    \begin{tabularx}{.8\textwidth}{ccccccc}
        \toprule
        \multicolumn{1}{l}{Batch Size} & Formation Energy & Bandgap & Density & E Fermi & EPA & A Density \\ \hline
        64   & .045 & .277 & .073 & .490 & 1.96 & .218 \\
        128  & \textbf{.043} & \underline{.271} & \underline{.070} & \underline{.485} & \underline{1.85} & \textbf{.213} \\
        256  & \underline{.044} & \textbf{.270} & \textbf{.068} & \textbf{.479} & \textbf{1.82} & \underline{.214} \\
        \bottomrule
    \end{tabularx}
    \caption*{Comparative results showing the effect of batch size in predicting MAE for six material properties. The best values are in bold, the second-best are underlined.}
    \label{tab:batchsize_effect}
    
\end{table}
%%%%%%%%%%%%%%%%%%%%%%%%%%%%%%%%%%%%%%%%
\begin{table}[!t]
    \centering
    
    \captionof{table}{The effect of changing the configuration of supervised Barlow-Twin on property prediction task.}
    \renewcommand{\arraystretch}{1.5}
    \setlength{\tabcolsep}{7pt}
    \begin{tabularx}{.8\textwidth}{ccccccc}
        \toprule
        \multicolumn{1}{l}{Loss Type} & Formation Energy & Bandgap & Density & E Fermi & EPA & A Density \\ \hline
        On diagonal only   & \underline{.045} & \underline{.282} & .079 & \underline{.504} & 1.96 & \underline{.225} \\
        Off diagonal only  & \textbf{.043} & .285 & \underline{.075} & .508 & \underline{1.92} & .229 \\
        SPMat-BT   & \textbf{.043} & \textbf{.271} & \textbf{.070} & \textbf{.485} & \textbf{1.85} & \textbf{.213} \\
        % 256  & \underline{.044} & \underline{.270} & .068 & \underline{.479} & \underline{1.82} & \underline{.214} \\
        \bottomrule
    \end{tabularx}
    \caption*{Comparative results are showing the effect of using standalone on-diagonal element and off-diagonal element of the supervised Barlow-Twin loss in predicting MAE for six material properties. The best results are from SPMat-BT and made bold, while the second best is underlined.}
    \label{tab:off_on_diag}
\end{table}
%%%%%%%%%%%%%%%%%%%%%%%%%%%%%%%%%%%%%%%
\begin{table}[!t]
    \centering
    
    \captionof{table}{The effect of using surrogate labels in SSL pretraining on classification tasks.}
    \renewcommand{\arraystretch}{1.5}
    \setlength{\tabcolsep}{7pt}
    \begin{tabularx}{.45\textwidth}{ccccccc}
        \toprule
        \multicolumn{1}{l}{Method} & IS-GAP& IS-METAL & IS-MAGNETIC \\ \hline
        BT  & \underline{.8781} & .9174 & \underline{.9377} &  \\
        SC  & .8717 & \underline{.9267} & .9281  \\
        SPMat  & \textbf{.8956} & \textbf{.9384} & \textbf{.9529}  \\
        % 256  & \underline{.044} & \underline{.270} & .068 & \underline{.479} & \underline{1.82} & \underline{.214} \\
        \bottomrule
    \end{tabularx}
    \caption*{Comparative results are showing the effect of using supervised pretraining on the downstream classification task for metallicity, magnetic behavior, and type of band gap. The best values are in bold and second best underlined.}
    \label{tab:class}
\end{table}
%%%%%%%%%%%%%%%%%%%%%%%%%%%%%%%%%%%%%%%%%%%%%%%5
\subsection*{Effect of different training configurations for SPMat-BT}
Since the objective function of the proposed SPMat-BT (Eqn.~\ref{simple-BT}) comprises two different terms—where the first term represents the correlation term (for materials within the same class) and the second represents the decorrelation term (for materials from different classes)—we aim to understand their individual effects on pretraining and downstream performance.
To this end, we pretrain our setup using one term at a time and report the results in Table~\ref{tab:off_on_diag}. As shown, the results across all properties indicate that the proposed combination produces the best performance, demonstrating the importance of SPMat-BT.
The t-SNE analysis further supports this observation: compared to the full SPMat-BT model shown in Fig. \ref{tsne}, the t-SNE plots from the individual components in Fig. \ref{tsne2}, while showing some degree of separability, do not exhibit as strong separability as the complete SPMat-BT configuration.

\subsection*{Effect of supervised pretraining on classification performance}
% We test the efficacy of supervised pretraining on the classification performance for three binary classification tasks, such as IS-GAP-DIRECT, IS-METAL, and IS-MAGNETIC, from the MP dataset in Table \ref{tab:class}. We compare the classification accuracy of SSL-BT and SSL-SC with SPMat-BT pretrained with bandgap as surrogate labels. It is seen that in the case of IS-GAP-DIRECT and IS-MAGNTIC classification tasks, the SPMat-BT (our proposed) has a clear advantage over SSL-BT and SSL-SC setups. In the case of the IS-METAL classification task, our proposed model performs slightly better than the SSL models. Notably, the IS-GAP classification task determines whether the band gap is direct, the IS-METAL classification task determines whether the material is a metal, and the IS-MAGNETIC classification task determines the magnetic characteristics of the material. Therefore, the improvement in classification performance demonstrates that the guided pretraining using surrogate labels learns a better representation in comparison to the unguided SSL pretraining.
We evaluate the efficacy of supervised pretraining on three binary classification tasks—IS-GAP-DIRECT, IS-METAL, and IS-MAGNETIC—from the MP dataset (Table~\ref{tab:class}). We compare the classification accuracy of SSL-BT and SSL-SC with SPMat-BT pretrained using bandgap as surrogate labels. 
In all three classification tasks, SPMat-BT outperforms SSL-BT and SSL-SC, showing a 2.00\% improvement for IS-GAP-DIRECT, a 1.62\% improvement for IS-MAGNETIC, and a 1.26\% improvement for IS-METAL over the second-best model.
These results demonstrate that guided pretraining with surrogate labels enables better representation learning compared to unguided SSL pretraining.

%%%%%%%%%%%%%%%%%%%%%%%%%
%%%%%%%%%%%%%%%%%%%%%%%%%%%%%%
%%%%%%%%%%%%%%%%%%%%%%%%%%%%%%%%%%%%%%%%
\section*{DISCUSSION}
%The Discussion should be succinct and must not contain subheadings.

As the demand for advanced materials continues to increase, accurate property prediction is becoming increasingly crucial for driving material discovery and innovation. In this regard, AI methods have shown promise but typically require extensive datasets to perform well for each property. Instead, with the rise of self-supervised learning (SSL) approaches, there is potential to develop a \textit{foundation} model where a single model captures significant information and knowledge, which can later be fine-tuned for different property prediction tasks.
We advocate for a research shift toward integrating SSL strategies to develop foundation models that capture broad and transferable knowledge, enabling accurate prediction and deeper understanding across a wide range of material properties.
In this study, we introduced SPMat, a novel SSL framework for material science that employs surrogate label based supervised pretraining to improve material property prediction. First, we developed novel pretext objective functions for two popular SSL frameworks, adapting them to accommodate labels for supervised pretraining. Second, to enhance the diversity of augmentations, we introduced GNDN, a novel augmentation technique designed to improve model learning by introducing controlled variations in neighbor distances within graph representations.
We extensively evaluated SPMat in the Materials Project database on six distinct material properties. Our results demonstrated significant improvements over baseline methods, showcasing the effectiveness of surrogate label driven supervised pretraining and augmentation strategies for material property prediction. SPMat consistently achieved superior performance, underlining its potential to accelerate material discovery and understanding.

In this study, we primarily compared SPMat with two types of approach: existing deep learning methods trained end-to-end to predict specific properties and more recent SSL approaches, where pre-training is conducted first, followed by fine-tuning for individual property prediction tasks. Although SPMat is computationally expensive to train compared to traditional deep learning methods, its computational requirements are comparable to those of SSL-based approaches. Importantly, during inference, the computational requirements for existing deep learning methods, SSL approaches, and SPMat are identical. However, as demonstrated earlier, SPMat achieves significantly better generalization across properties.
It should be noted that SPMat leverages surrogate labels during pre-training, which could be considered a limitation due to the dependence on their availability. However, with large databases such as the Materials Project, the availability of surrogate labels is becoming less of a challenge. This makes it feasible to train foundation models like SPMat, especially when they provide significant improvements in property prediction and generalization.

Even though SPMat's performance gains in MAE across benchmarks range from 2\% to 6.67\%, they have important chemical and practical implications for material property prediction. Small reductions in prediction error have a significant influence on high-throughput screening and material discovery in materials science, where precise candidate identification is essential for applications like photovoltaics, batteries, and catalysis. For example, a 5\% MAE improvement in bandgap prediction increases the accuracy of choosing materials with particular electronic characteristics, such as those with bandgaps in the 1.5-2 eV range that are ideal for solar cell applications. Stable or high-performing materials for battery electrodes or catalysts can be prioritized through more accurate predictions of formation energy or conductivity, which also lowers the quantity of false positives and negatives in screening processes. In summary, the 2–6.67\% MAE improvements achieved by SPMat result in enhanced precision in material screening, reduced experimental costs, and better capture of chemically relevant structural-property relationships. These outcomes highlight the practical and chemical significance of our approach.  

This study has certain limitations that pave the way for future research. First, we did not explore the impact of different backbone architectures, including more advanced designs such as transformers~\cite{vaswani2017attention}. Given their proven effectiveness in various domains, we anticipate that adopting such architectures could further enhance the performance of SPMat. Investigating the influence of different backbones should be addressed in future studies to fully optimize the framework's potential.
Furthermore, future work could expand on this study by evaluating the transferability of SPMat, particularly for smaller-dimensional materials such as 2D and 1D systems. Intuitively, foundational models such as SPMat, which demonstrated promising results in this study, are expected to capture critical information that can be generalized well to these lower-dimensional systems. However, material science applications may require a more in-depth analysis to address the unique challenges and intricacies of transfer learning in this context.
Despite these limitations, SPMat provides a robust foundation and strong motivation for further exploration in these directions. Using its ability to generalize across diverse material properties, future studies can build on this framework to unlock new insights and advance the development of property prediction models for a broader range of material systems.

\section*{METHODS}
This section discusses the methodology of our proposed SSL-based foundation model pre-training and fine-tuning for material property prediction. The main target of the proposed scheme is to learn robust representation from the 
%un-labeled and 
labeled crystallographic input data. In the case of crystal structures of materials, we create dissimilar views of a single crystal structure of a particular material by adopting carefully chosen augmentation on the actual structure. The label accommodated SimCLR or Barlow-Twin contrastive loss is applied to these different views of the same data point termed as positive pairs, and the loss objectives try to bring the augmented versions of the positive pairs closer in the embedding domain or increase correlation depending on the loss objective being applied. It simultaneously tries to keep the embedding of other crystal structures or negative samples apart or reduce correlation. These increases or decreases in correlation or distance depend on the class information collected from the surrogate labels. Therefore, samples from the same class data point will either be closer to each other or their correlation will be maximized in comparison to the samples from different class data points. To leverage label information in a supervised SPMat SSL setup, we introduce these class labels to associate with specific materials. For example, we use labels such as whether a material is metallic or non-metallic and magnetic or non-magnetic during the pretraining phase. This approach helps ensure that materials with similar characteristics are positioned closer together in the embedding space. These labels are termed "surrogate labels" because the downstream task is regression, and these labels primarily serve to cluster materials with similar broad characteristics rather than directly inform the regression task. During pretraining the foundation model, as previously mentioned we utilize a label incorporated SimCLR-based contrastive loss that pulls embeddings from the same class closer together and pushes embeddings from different classes farther apart within a mini-batch. Furthermore, we propose a novel contrastive Barlow-Twin loss that incorporates class information into SSL pretraining. The standard Barlow-Twin loss aims to produce an empirical cross-correlation matrix by comparing the embeddings of different data points. Ideally, by the end of pretraining, the diagonal elements of this matrix, representing the correlation between different views of the same data point, should be close to 1, indicating high similarity. Conversely, the off-diagonal elements, which compare embeddings from different data points, should approach zero, indicating minimal correlation. In our proposed approach, class information is integrated into the loss function. This means that the empirical cross-correlation matrix is adjusted to ensure that off-diagonal elements—representing data points from the same class—approach 1. This adjustment effectively draws embeddings of data points from the same class closer together in the embedding space, even if they represent different data points. This modified loss function enables the model to learn more refined representations of material structures by capturing intra-class similarities and inter-class differences during the SSL pretraining phase. Later, the models pre-trained with un-supervised and supervised contrastive setups are finetuned and tested for the regression task. To achieve variation in the crystal structures of a material, three types of augmentation techniques, such as atom masking, edge masking, and random noising of the neighbor distance on the graph (GNDN), have been used. The GCNN-based encoder is used and pre-trained with contrastive un-supervised and supervised loss functions driven by surrogate labels to get the embeddings from material structures. This pre-trained foundation model is utilized later for fine-tuning the downstream tasks of material property prediction. 

\subsection*{Graph level neighbor distance noising}
% Here we discuss the technical details of GNDN.
GNDN augments crystal graphs directly by perturbing neighbor distances to enhance the resilience of SSL pretraining. For a crystal structure loaded from a CIF file and for each atom $i$ in that crystal structure we compute the distances $d_{ij}$ to the $M$ nearest neighbors within an R~\AA{} radius, accounting for periodic boundary conditions. Noise $\epsilon_{ij} \sim U(-\delta, \delta)$~\AA{}ngstroms is sampled independently for each distance and added element-wise, perturbing each distance as $d_{ij}' = d_{ij} + \epsilon_{ij}$.
% \[
% d_{ij}' = d_{ij} + \epsilon_{ij}. 
% \]
% The perturbed distances are expanded using a Gaussian basis:
% \[
% f_k(d_{ij}') = \exp\left(-\frac{(d_{ij}' - \mu_k)^2}{\sigma^2}\right),
% \]
%with $\mu_k = 0, 0.2, \dots, 8.0$ and $\sigma = 0.2$~\AA{}ngstroms.
These perturbed distances are expanded into a feature vector using a Gaussian basis with centers $\mu_k$ ranging from 0 to 8~\AA{}ngstroms in steps of 0.2~\AA{}ngstroms, yielding $f_k(d_{ij}') = \exp\left(-\frac{(d_{ij}' - \mu_k)^2}{\sigma^2}\right)$, where $\sigma = 0.2$~\AA{}ngstroms. This process generates edge features for two augmented views of the same crystal, which are fed into the graph neural network for SSL pretraining.
As the process is repeated for two views of the same crystal, generating distinct edge features for SSL is attained. The noise range reflects typical atomic displacements (e.g., thermal vibrations), while the uniform distribution ensures balanced perturbations. If $d_{ij}'$ becomes negative (rare for $d_{ij} > 0.5$~\AA), the Gaussian expansion accommodates it without clipping. GNDN complements atom masking and edge masking, forming a robust augmentation pipeline.

\begin{algorithm}
\caption{Graph-level Neighbor Distance Noising (GNDN)}
\begin{algorithmic}[1]
\State \textbf{Input}: A crystal graph $G = (V, E)$, where $V = \{v_1, v_2, \ldots, v_n\}$ is the set of nodes (atoms) and $E$ is the set of edges (connections between neighboring atoms), with each edge $e_{ij} \in E$ associated with a distance $d_{ij} = ||\mathbf{r}_i - \mathbf{r}_j||$ (where $\mathbf{r}_i$ and $\mathbf{r}_j$ are position vectors of atoms $v_i$ and $v_j$), and a noise magnitude parameter $\delta$.
\State \textbf{Output}: Augmented edge features $F$ for a single view of the graph.
\For{each set of edge distances $\{d_{ij} \mid e_{ij} \in E\}$}
    %\State Sample noise $\epsilon_{ij} \sim \mathcal{U}(-\delta, \delta)$ for each edge $e_{ij}$
    \State Sample noise $\epsilon_{ij} \sim \mathcal{U}(-\delta, \delta)$ for each edge $e_{ij}$ \Comment{$\delta$ controls the noise magnitude in \AA{}ngstroms}
    \State Perturb distances $d'_{ij} = d_{ij} + \epsilon_{ij}$ \Comment{Apply GNDN}
    \State Compute Gaussian expansion $F = \text{GaussianExpand}(\{d'_{ij}\})$
    \State \Return $F$
\EndFor
\end{algorithmic}
\end{algorithm}

\subsection*{GCNN based encoder}
Selecting a suitable encoder to create embedding for SSL pretraining is crucial while designing such schemes. Graph Neural Networks (GNNs) are appropriate tools to map crystalline structures in graphs and then create embeddings from that crystal structure by convolution for further operations. In our supervised pretraining scheme for contrastive SSLs, we select the Crystal Graph Convolutional Neural Network (CGCNN)~\cite{xie2018crystal} based encoder because it is specifically designed to handle the unique characteristics of crystal structures, where atoms are arranged in a repeating, three-dimensional lattice. It represents these structures as graphs, with atoms as nodes and bonds as edges, naturally capturing the spatial relationships and interactions within a crystal. This graph-based representation allows CGCNN to effectively model atoms' local and global connectivity, which is crucial for understanding the material’s properties.

CGCNN uses graph convolutional operations that aggregate information from the neighbors of an atom, allowing it to learn complex interactions between atoms. By learning these interactions, CGCNN can generate detailed and informative embeddings of the material structures, which can be used to predict various material properties. CGCNN's design makes it preferable for SSL because it can learn meaningful representations of material structures even with unlabeled data. The network can be trained to maximize similarities between different views of the same material structure while minimizing similarities between different materials, hence learning valuable features that can be transferred to downstream tasks like property prediction.

In the CGCNN encoder, atoms are considered nodes, and the interaction among atoms is called bonds. Bonds with neighboring atoms for a particular atom create a graph network structure and feature the convolution of an atom. Its surrounding atom updates the feature vector of an atom in different layers. 
\begin{equation}
v^{(t+1)}_i = g \Bigg[ \left( \sum_{j,k} v^{(t)}_j \oplus u_{(i,j)_k} \right) W^{(t)}_c + v^{(t)}_i W^{(t)}_s + %\\
b^{(t)}_c + b^{(t)}_s \Bigg]
\label{eq_cgcnn}
\end{equation}
In Eq.~\eqref{eq_cgcnn} $v^{(t+1)}_i$ is the updated feature vector of atom $i$ at $t+1$ iteration after $t^{th}$ layer of graph convolution, $g$ is a non-linearity, $v_j$ is the feature vector of nodes surrounding the node $i$, $u_{(i,j)_k}$ is the feature vector of $k^{th}$ bond between nodes $i \text{and} j$, $\oplus$ represents feature concatenation. Weight matrices are denoted as $W^{(t)}_c, \text{and}  ~W^{(t)}_s$. While, $b^{(t)}_c ,  \text{and}  ~b^{(t)}_s$ represents bias matrices. However, the Eq.~\eqref{eq_cgcnn} incorporate a weight matrix that assigns the same weight to all the neighbor irrespective of their influence on the center atom, and thus, the Eq.~\eqref{eq_cgcnn} is updated in Eq.~\eqref{eq2_cgcnn}. 

\begin{equation}
v^{(t+1)}_i = v^{(t)}_i + \sum_{j,k} \sigma\left(z^{(t)}_{(i,j)_k}\right) W^{(t)}_c + b^{(t)}_c \odot g\left(z^{(t)}_{(i,j)_k} W^{(t)}_s + b^{(t)}_s\right)
\label{eq2_cgcnn}
\end{equation}
In this context, $z^{(t)}_{(i,j)_k} = v^{(t)}_i \oplus v^{(t)}j \oplus u{(i,j)_k}$, where $\sigma(\cdot)$ is a weight matrix designed to learn and distribute weights to neighboring atoms according to their bond strengths with the central atom. The symbol $\odot$ represents the element-wise multiplication operation. After updating the feature vectors with the convolution operation, a pooling layer extracts the information about the overall crystal system. It creates a latent representation of the desired dimension as in Eq.~\eqref{eq3_cgcnn}. 
\begin{equation}
\mathbf{v}_c = \text{Pool}(\mathbf{v}_0^{(0)}, \mathbf{v}_1^{(0)}, \ldots, \mathbf{v}_N^{(0)}, \ldots, \mathbf{v}_N^{(R)})
\label{eq3_cgcnn}
\end{equation}
 Following the GNN encoder, we attach a projection head consisting of two fully connected layers (MLP) to create the final embedding. This embedding is applied to the self-supervised loss functions during the pretraining phase.

 To create a suitable augmentation for generating different views of material, and crystalline structures, we follow already established literature like AugLiChem~\cite{magar2022auglichem} and Crystal-Twins~\cite{magar2022crystal}. We utilized the Atom Masking (AM) and Edge Masking (EM) techniques sequentially following~\cite{magar2022crystal}. In AM, we randomly select 10\% of the atoms in the crystal structure and mask them by setting their feature vectors to zero. This simulates scenarios where certain atoms are missing or their contributions are obscured, either due to imperfections in the material or experimental limitations. Like atom masking, the EM technique involves randomly masking 10\%  of the edges (connections) between atoms in the crystal structure. Masking edges simulate the absence of specific bonding interactions or disruptions in the crystal's connectivity, which can occur in disordered or defective materials. 
 
 After applying AM and EM augmentation to the material structure, the neighborhood is selected, and a graph network is created for the particular material structure. Now at this stage, the GNDN applies uniform random noise to the neighbor distances directly with a range of $-.5$ to $.5$. This noise is applied to the distance features that define the graph's edges, representing the connections between atoms. Applying perturbation to the material structure directly may hamper the structure and produce physically implausible configurations, such as overlapping atoms or broken chemical bonds, which do not correspond to any actual material state. By introducing noise directly to the graph representation, we avoid creating unrealistic structures and instead focus on enhancing the model's ability to handle variability in the graph, which is a more controlled and meaningful augmentation. 

\subsection*{Self supervised contrastive setup}

In the self-supervised contrastive setup, a data point is augmented using random atom masking, random edge masking, or, in some cases, applying the proposed GNDN technique, and two versions of the same data point are generated. Then, contrastive loss is applied to increase the closeness of these two different views of the same data point in the latent space, and that is how the model learns the structure of materials. This simple architecture to learn data representation is already popular in computer vision and is named SimCLR (Simple framework for contrastive learning). If $x_{i}$ and $x_{j(i)}$ are the two augmented versions of the data point $x$ and $z_{i},z_{j(i)}$ are their embeddings created from passing the augmented structures through GNN encoder, pooling layers and two MLPs as projection head with non-linearity. Because of its simplicity and wide use, we intentionally use CGCNN as our GNN-based encoder to create embeddings. And the contrastive loss is applied on the embeddings $z_{i}$ and $z_{j(i)}$ batch-wise. 
% The loss function is designed in a way that it tries to find out $x_{j(i)}$ from $\{{x}_k\}_{k \neq i}$, where $\{x_k\}$ is a set of possible augmented versions and $x_i, x_{j(i)}$ are positive pairs. If the total minibatch during pre-training has N samples then the augmented version will create 2N samples out of it. Therefore, for a particular structure if there are two augmented views then in the mini-batch the other $2N-2$ structures are from other data points. Then the contrastive loss function for the augmented views of the same structure or positive pair is given as in eq~\eqref{eq1}. Here, $\mathbb{1}_{[k \neq i]} \in {0,1}$  is an indicator function that evaluates to 1 if the condition inside the brackets is true and 0 otherwise.  In eq~\eqref{eq1}, $\mathbb{1}_{[k \neq i]}$ ensures that when summing over all possible pairs, the term where $k = i$  is excluded to avoid comparing samples with itself. 

To express the relation mathematically, let us consider a set of \( N \) randomly sampled data-label pairs, \( \{ \mathbf{x}_k, \mathbf{y}_k \}_{k=1...N} \), The corresponding training batch consists of \( 2N \) pairs, \( \{ \tilde{\mathbf{x}}_\ell, \tilde{\mathbf{y}}_\ell \}_{\ell=1...2N}^{} \), where \( \tilde{\mathbf{x}}_{2k} \) and \( \tilde{\mathbf{x}}_{2k-1} \) represent two random augmentations (also referred to as "views") of \( \mathbf{x}_k \) for \( k = 1, \ldots, N \), and \( \tilde{\mathbf{y}}_{2k-1} = \tilde{\mathbf{y}}_{2k} = \mathbf{y}_k \) are their corresponding classes. Here, \( N \) samples are considered as a "batch," and the set of \( 2N \) augmented samples as a multi-viewed batch. If \( i \in I \equiv \{ 1 \ldots 2N \} \)
is the index of an arbitrary augmented sample, and $j(i)$ is the index of another augmented version of the same sample, then in \eqref{eq1} $A(i)$ represents all the samples except the anchor $i$, mathematically, $A(i) \equiv I\setminus i$. Index $j(i)$ is also termed as the positive of the anchor samples. The rest of the $2N-2$ indices are of negative samples and mathematically written as, $ k \in A(i) \setminus \{ j(i) \} $
% \begin{equation}
% \mathcal{L}_{i,j} = - \log \frac{\exp(\text{sim}(z_i, z_j) / \tau)}{\sum_{k=1}^{2N} ~
% \mathbb{1}_{[k \neq i]} ~\exp(\text{sim}(z_i, z_k) / \tau)}
% \end{equation}

\begin{equation}
\mathcal{L}^{\text{self}} = \sum_{i \in I} \mathcal{L}^{\text{self}}_i = - \sum_{i \in I} \log \frac{\exp \left( z_i \cdot z_{j(i)} / \tau \right)}{\sum_{a \in A(i)} \exp \left( z_i \cdot z_a / \tau \right)}
\label{eq1}
\end{equation}
Here, $z_i$ and $z_{j(i)}$ originate from the same data sample hence termed as positives.
%They actually are members of a set of embeddings ${z}_{\ell} = \text{Proj}(\text{Enc}(\tilde{\mathbf{x}}_{\ell})) \in \mathbb{R}^{D_p}created from all data points.  
The temperature parameter $\tau$ scales the similarities before applying the exponential function. It controls the sharpness of the similarity distribution. Lower values of $\tau$ make the model more sensitive to differences in similarity scores, while higher values make the distribution smoother. The final loss is computed across all positive pairs in both directions, meaning that the loss for pair $(i, j)$ and the loss for pair $(j, i)$ are included. This ensures symmetry in the loss computation. Here, $(\cdot)$ represents the cosine similarity among two vectors in \eqref{eq1}. In the SimCLR framework, the contrastive loss only considers positive pairs derived from augmented views of the same sample. It treats all other samples in the batch as negatives, regardless of their class. This approach does not utilize any available class label information, which limits the ability of the model to learn sophisticated distinctions between samples that belong to the same class versus those that do not. Furthermore, the performance of SimCLR is highly dependent on the batch size because it relies on a large number of negative samples to learn representations effectively. Larger batch sizes provide more negative examples, improving the contrastive learning process but also requiring more computational resources and memory.

% \begin{equation}
% \text{sim}(\mathbf{u}, \mathbf{v}) = \frac{\mathbf{u}^\top \mathbf{v}}{\|\mathbf{u}\| \|\mathbf{v}\|}
% \label{eq2}
% \end{equation}

\subsection*{Supervised contrastive setup driven by surrogate labels}
The recently developed Supervised version of Contrastive (SupCon) loss takes advantage of class labels by allowing multiple positive pairs (samples from the same class) and multiple opposing pairs (samples from different classes) for each anchor. The SupCon loss, however, utilizes all samples of the same class as positive pairs, significantly improving the learning of class-specific features. While SupCon loss also benefits from larger batch sizes, using multiple positive pairs reduces its dependence on batch size. The model can achieve effective learning even with smaller batch sizes by using class labels, making it more flexible and efficient. To leverage from the available labels, supervised setup in contrastive self-supervised learning is a popular and novel advancement. A fully supervised self-supervised batch contrastive technique has already been proposed for image object classification problems and claims to achieve better performance. Therefore, for the downstream regression task of material property prediction, we aim to incorporate material class information as labels into the loss function to analyze the effect of introducing surrogate labels-driven supervision on the material property prediction task. To achieve that, we explore a broader classification of materials based on their magnetic, metallic, and bandgap-related classes available in the MP database. A similar technique used in~\cite{khosla2020supervised} has been adopted to design the loss function and to inject class labels that is termed as surrogate labels for downstream regression tasks. The pre-task is essentially the same as un-supervised contrastive learning except for the supervised case, the positive samples are from the same class and pulled closer on the contrary the negative samples from different classes are pulled apart. The technical advantage of this process is that the number of positive samples increases in a mini-batch, which is balanced with the number of negative samples. 
\begin{equation}
\mathcal{L}_{\text{sup}}^{} = \sum_{i \in I} \mathcal{L}_{\text{sup},i}^{} = \sum_{i \in I} \frac{-1}{|P(i)|} \sum_{p \in P(i)} \log \frac{\exp \left( z_i \cdot z_p / \tau \right)}{\sum_{a \in A(i)} \exp \left( z_i \cdot z_a / \tau \right)}
\label{sup_con_eq}
\end{equation}
In implementing the loss function in Eq.~\eqref{sup_con_eq} we follow the procedure described in~\cite{khosla2020supervised}. Here, $\mathcal{L}_{\text{sup}}$ is the supervised contrastive loss computed over all samples in a batch, $P(i)$ denotes the set of indices that have the same class label as the anchor sample $i$ except for $i$ itself. $ \left|P(i) \right|$  is the number of positive samples for the anchor sample $i$, $A(i)$ denotes the set of all possible indices except the anchor sample $i$, including both positive and negative samples, $z_p$ is the embedding of a positive sample $p$ that shares the same class label as the anchor sample $i$ and $z_a$ is the embedding of any sample $a$ in the set $A(i)$. 
\subsection*{Self-supervised Barlow Twin setup}
Barlow Twins is an efficient self-supervised learning framework introduced to address the challenge of redundancy in learned representations. It was first presented in~\cite{barlow1961possible, barlow2001redundancy} and later introduced in the SSL framework by Zbontar~et al.\cite{zbontar2021barlow}. The core idea behind Barlow Twins is to ensure that the embeddings of augmented views of the same image (or data point) are highly similar while minimizing the redundancy between different dimensions of these embeddings. This is achieved through a novel objective function that targets the cross-correlation matrix of the embeddings obtained from two different views of the same batch of data. The model takes two augmented views of each input data point. These augmentations are typically random transformations applied to the original data, such as atom masking, edge masking, random perturbation, etc. For our SSL, following the previous literature~\cite{magar2022crystal}, we use CGCNN for material structure data to process each augmented view to produce embeddings. The same network (with shared weights) is used for both views to ensure consistency. The embeddings from the two augmented views are used to compute an empirical cross-correlation matrix. This matrix measures the similarity between the corresponding dimensions of the two sets of embeddings. Given two sets of embeddings with dimension $D$ as $z_1$, $z_2$ $\in \mathbb{R}^{B\times D}$ obtained from augmented views of the same data batch $N$, the cross-correlation matrix $C$ as in~\eqref{BT-1} measures the similarity between the embeddings from these two views. The similarity is increased if they are from the same feature dimension across views and decreased if they are from different feature dimensions. Therefore, the empirical cross-correlation matrix ideally becomes an identity matrix with diagonal elements '1' and off-diagonal elements '0'. The embeddings are normalized to have zero mean and unit variance. In~\eqref{BT-1} $\tilde{z}_{1i}^{(b)}$ represents the $i^{th}$ component of the normalized embedding of the $b^{th}$ sample in $z_1$ and $\tilde{z}_{2j}^{(b)}$ represents the $j^{th}$ component of the normalized embedding of the $b^{th}$ sample in $z_2$. Most importantly, the Barlow Twin loss aims to learn representations invariant to input distortions while minimizing redundancy between feature dimensions. It achieves this by ensuring the cross-correlation matrix between embeddings of two views of the same data point is close to the identity matrix. This encourages each feature in one view to correlate only with the corresponding feature in the other view, promoting non-redundant and informative representations.
%%%
\begin{equation}
C_{ij} = \frac{1}{N} \sum_{b=1}^{N} \tilde{z}_{1i}^{(b)} \tilde{z}_{2j}^{(b)}
\label{BT-1}
\end{equation}
The cross-correlation matrix can be written in matrix form as in~\eqref{BT-2}.
Where $\tilde{z}_1$ and $\tilde{z}_2$ are the matrices of normalized embeddings for the two views and $\tilde{z}_1^T$ is the transpose of  $\tilde{z}_1$. 
\begin{equation}
C = \frac{1}{N} \tilde{z}_1^T \tilde{z}_2
\label{BT-2}
\end{equation}
%%%%%%%%%%%%%%
\begin{equation}
\mathcal{L}_{\text{BT}} = \sum_{i} (1 - C_{ii})^2 + \lambda \sum_{i} \sum_{j \neq i} C_{ij}^2
\label{BT-3}
\end{equation}

\subsection*{Contrastive supervised Barlow Twin setup}
Our careful observation reveals that the original Barlow Twins framework considers only the embeddings of augmented samples from the same data point when calculating the off and on-diagonal terms of the empirical cross-correlation matrix. The loss function is designed to maximize the correlation between identical embeddings of the augmented views of each data point. In contrast, the off-diagonal terms compute the correlation between different embedding dimensions across the distinct views of the data point, and the Barlow loss seeks to increase their dissimilarity by driving these correlations closer to zero. Therefore, the diagonal terms represent the correlation between the same features across both views, averaged over all samples, and the Barlow Twin loss maximizes these to enforce invariance between views of the same data points. However, the off-diagonal terms capture correlations between different feature dimensions across the batch, not between different data points, and are minimized to reduce feature redundancy. These terms of cross-correlation matrix do not directly compare embeddings of different samples. However, if the class information is available as labels, it can contribute to a better representation learning and may generate improved results in downstream tasks. Therefore, we re-designed the original Barlow-Twin in a novel way so that it can consider embeddings from the same class as a positive sample and try to increase their correlation. Hence, in our supervised Barlow setup not only the on-diagonal elements but also the off-diagonal elements having the same class are considered as positive and their correlation is maximized. Hence, in our Supervised Barlow Twins framework, we replace the feature-wise similarity matrix of the original Barlow-Twin with a sample-wise similarity matrix. In our Supervised Barlow Twin framework, we redefine $C = \frac{1}{D}~\tilde{z}_{1} \tilde{z}^{T}_{2} \in \mathbb{R}^{N\times N}$ as a sample-wise similarity matrix, where $C_{ij}$ measures the similarity between samples $i$ and $j$ across views. 

\begin{equation}
\mathcal{L}_{\text{same-class}} = \sum_{i,j} M_{ij} (1 - C_{ij})^2
\label{bt_sup1}
\end{equation}
%%%%%%%%%
\begin{equation}
\left\{\begin{matrix}
M_{ij} = 1 &\text{if}  &y_i = y_j  &\text{ the samples belong to the same class)} \\ 
M_{ij} = 0 &\text{if}  &y_i \neq  y_j   &\text{ the samples belong to the different class)}  
\end{matrix}\right.    
\label{bt_sup3}
\end{equation}
In \eqref{bt_sup1} multiplying with the mask $M_{ij}$  defined in \eqref{bt_sup3} and created from surrogate labels $y$ effectively makes the on-diagonal and some of the off-diagonal elements of the empirical cross-correlation matrix $1$ and they are considered as same class while calculating the loss $\mathcal{L}_{\text{same-class}}$. This approach ensures that the embeddings of the same class are similar, including both strict diagonal and same-class off-diagonal elements. To design the loss function for the different class samples, we follow the approach described in~\cite{tsai2021note}. The modified loss for dissimilar samples tries to make them as dissimilar as possible, contrary to the original Barlow Twin implementation. Also, it considers only the samples with different classes as its candidate if it is an off-diagonal element in the correlation matrix and not of the same class. 
\begin{equation}
\mathcal{L}_{\text{diff-class}} = \lambda \sum_{i \ne j} (1 - M_{ij})(1 + C_{ij})^2
\label{bt_sup2}
\end{equation}
\begin{equation}
\mathcal{L}_{\text{Sup-BT}} =\sum_{i,j} M_{ij} (1 - C_{ij})^2 + \lambda \sum_{i \ne j} (1 - M_{ij})(1 + C_{ij})^2
\label{bt_sup3}
\end{equation}
Minimizing the total loss function in the supervised Barlow Twins framework works by adjusting the model parameters to reduce both the same and the different-class loss. Empirical Cross-Correlation Matrix $C$ is computed from the model's output embeddings for two augmented views of the input batch. It captures the pairwise similarities between the embeddings. For same-class pairs $M_{i,j} = 1$, the term $(1 - C_{ij})^2$ is minimized when $C_{ij}$ is close to 1. This means embeddings of samples from the same class should be as similar as possible. For different-class pairs $where, M_{i,j} = 0$, the term $(1 + C_{ij})^2$ is minimized when $C_{ij})$ is close to -1. This means embeddings of samples from different classes should be as dissimilar as possible. This essentially resembles supervised contrastive loss but uses a squared penalty and operates on the full similarity matrix rather than a softmax-normalized form. This makes it computationally efficient avoiding the need for temperature scaling or log-exp operations.
The total loss combines these two objectives, balancing them using the weight $\lambda$. 

From a machine learning point of view, the inclusion of same-class information improves representation learning by encouraging intra-class similarity. The model learns to group these points closer in the feature space by considering different data points within the same class as positives. This encourages intra-class similarity, making the embeddings of data points from the same class more similar. At the same time, different pairs of classes are pushed apart, ensuring that the embeddings of different classes are dissimilar. This enhances the discriminative power of the learned embeddings. Using label information explicitly, instead of unsupervised methods that rely solely on data augmentations, leverages the available supervised signal to improve the quality of the learned representations. The novel supervised Barlow-Twins loss function is designed to ensure that embeddings of the same class are similar (both on-diagonal and off-diagonal elements) and embeddings of different classes are dissimilar. In material property prediction, embeddings must capture diverse structural or compositional features to predict properties like strength or conductivity. Our proposed loss prevents dimensional collapse by enforcing high similarity among the same class samples and dissimilarity for different class samples. This dual constraint ensures that embeddings span a space sufficient to distinguish property classes, avoiding collapse into a trivial solution, as any such collapse would violate both terms’ optimization goals. Unlike Barlow Twins’ feature-redundancy focus, our class-driven regularization is particularly suited to materials data, where property-specific diversity is critical.

The parameter $\lambda$ in the loss function is a weighting factor to balance the contributions of the loss of the same class and the loss of the different class. Without $\lambda$, the same class loss and the different class loss might have different magnitudes. This imbalance could cause the optimizer to focus more on minimizing one part of the loss at the expense of the other. Introducing $\lambda$ it is possible to adjust the influence of the loss of different classes, ensuring that both terms contribute appropriately to the total loss. If $\lambda$ is too high, the model might focus too much on separating different classes, potentially ignoring the intraclass similarity. If $\lambda$ is too low, the model may focus too much on clustering samples of the same class, potentially leading to less discriminative features between different classes.
\begin{algorithm}
\caption{Supervised Barlow-Twin Loss with Sample-Wise Cross-Correlation}
\begin{algorithmic}[1]
    \State \textbf{Input}: 
    \State \quad $z_1, z_2 \in \mathbb{R}^{B \times D}$ \Comment{Embeddings of two views, $B$ = batch size, $D$ = feature dimension}
    \State \quad $y \in \mathbb{R}^B$ \Comment{Class labels for the batch}
    \State \quad $\lambda$ \Comment{Weighting hyperparameter}
    \State \quad $D$ \Comment{Feature dimension}
    \State \textbf{Output}: 
    \State \quad $L$ \Comment{Supervised loss value}
    \Statex
    \State // Normalize embeddings separately (L2 normalization per sample)
    \State $\tilde{z}_1 \gets \text{F.normalize}(z_1, \text{dim}=1)$ %\Comment{$\tilde{z}_1[k, :] = z_1[k, :] / \| z_1[k, :] \|_2$}
    \State $\tilde{z}_2 \gets \text{F.normalize}(z_2, \text{dim}=1)$ %\Comment{$\tilde{z}_2[l, :] = z_2[l, :] / \| z_2[l, :] \|_2$}
    \Statex
    \State // Compute sample-wise cross-correlation matrix
    \State $S \gets (\tilde{z}_1 \times \tilde{z}_2^T) / D$  \Comment{Matrix multiplication,~$z_1$, $z_2$ $\in \mathbb{R}^{B\times D}$, $S \in \mathbb{R}^{B \times B}$}
    \Statex \Comment{$S_{kl}$ represents similarity between sample $k$ in $z_1$ and sample $l$ in $z_2$}
    \Statex
    \State // Create same-class mask based on labels
    \State $M \gets \text{zeros}(B, B)$  \Comment{Initialize $B \times B$ mask}
    \For{$k \gets 1$ to $B$}
        \For{$l \gets 1$ to $B$}
            \If{$y[k] = y[l]$}
                \State $M[k, l] \gets 1$  \Comment{Same-class pair}
            \Else
                \State $M[k, l] \gets 0$  \Comment{Different-class pair}
            \EndIf
        \EndFor
    \EndFor
    \Statex
    \State // Compute on-diagonal loss (same-class similarity)
    \State $on\_diag \gets 0$
    \For{$k \gets 1$ to $B$}
        \For{$l \gets 1$ to $B$}
            \If{$M[k, l] = 1$}
                \State $on\_diag \gets on\_diag + (1 - S[k, l])^2$
            \EndIf
        \EndFor
    \EndFor
    \Statex
    \State // Compute off-diagonal loss (different-class dissimilarity)
    \State $off\_diag \gets 0$
    \For{$k \gets 1$ to $B$}
        \For{$l \gets 1$ to $B$}
            \If{$M[k, l] = 0$}
                \State $off\_diag \gets off\_diag + (1 + S[k, l])^2$
            \EndIf
        \EndFor
    \EndFor
    \Statex
    \State // Combine losses with weighting factor $\lambda$
    \State $L \gets on\_diag + \lambda \times off\_diag$
    \State \Return $L$
\end{algorithmic}
\end{algorithm}
Our Supervised Sample-Wise Barlow Twins Loss offers obvious advantages over the original Barlow Twins, SimCLR, and Supervised SimCLR (SupCon). Compared to Barlow Twins, our approach utilizes labels to align same-class samples and separate different-class ones, making it directly suited for downstream tasks. Our loss uses labeled data to customize embeddings for particular tasks, whereas Barlow Twins, which lack label information, learn general-purpose representations that might not be intrinsically class discriminative. Moreover, Barlow Twins optimizes feature correlations to enforce view invariance and feature independence, which may scatter same-class samples if their features align differently across views. In contrast, our sample-wise focus optimizes sample similarities by directly clustering same-class samples in the embedding space. This guarantees that the class structure is more accurately reflected in our embedding. Furthermore, unlike Barlow Twins, which just eliminates feature duplication without addressing sample-level class differences, our loss clearly pushes different-class samples away, establishing sharper class borders that are crucial for supervised tasks. 

Compared to SimCLR, our loss removes the need for negative sampling, simplifying the training process. SimCLR relies on contrasting positive pairs against many negative pairs, which requires large batch sizes to ensure effective contrast. In contrast, our approach leverages all sample pairs within the batch, integrating both positive and negative relationships directly into the matrix, eliminating the need for explicit negative sampling. Moreover, our use of squared terms, such as $(1-C_{ij})^2$ provides smooth and gradual penalties, which can enhance optimization stability, particularly in the presence of noisy data. In contrast, SimCLR’s sharper log-softmax loss (InfoNCE), which relies on $\tau$ for temperature scaling, tends to be more sensitive to outliers and requires careful tuning of $\tau$ to achieve optimal performance.

Against Supervised SimCLR (SupCon), our loss presents a simpler formulation, utilizing direct squared penalties on all pairs in C without the need for a complex log-softmax or temperature parameter ($\tau$), which SupCon requires for tuning and normalization across similarities. This reduction in hyperparameters, just $\lambda$ versus $\tau$ and batch size considerations, makes our approach more straightforward to implement and tune. Furthermore, our approach uniquely blends Barlow Twins’ matrix concept with supervised contrastive goals, using a sample-wise cross-correlation matrix with absolute targets and squared penalties, distinct from SupCon’s relative, softmax-driven method.

%%%%%%%%%%%%%%%%%%%%%%%%%%%%training details%%%%%%%%%%%%%%%%%%%%
\subsection*{Training details}
Here, we discuss the details of the proposed SPMat training procedure for learning material representation and property prediction. The proposed architecture uses the graph encoder and associated fully connected layers to create and compare the embeddings against surrogate labels guided supervised contrastive Barlow-Twin loss and SimCLR loss for pretraining purposes. Hyper-parameters related to the graph formation and encoder are kept as prescribed in the original CGCNN paper~\cite{xie2018crystal}. The pretraining dataset was curated from the MP~\cite{jain2013commentary} database, and only material structures flagged as unstable were used in pretraining. As we injected class information as surrogate labels during pretraining, depending on the availability of labels, the number of pretraining data points might change slightly but never crossed 121371, which is the number of unstable structures available in the MP database when it was accessed. For both schemes, the pretraining was carried out for 15 epochs. The embedding dimension was kept at 128 for all the experiments. The models were pre-trained with the Adam optimizer, using a learning rate of 0.00001 and a weight decay of $1e^{-6}$ to prevent overfitting. We evaluated the performance of the model at the end of each epoch with a 95\%-5\% split.

The pretraining was carried out for the SPMat Barlow-Twin model using a batch size of 128. The loss function was tailored to the supervised Barlow-Twin approach, with an embedding size of 128 and a lambda regularization parameter set to 0.0051. The batch size for the loss function was 128, which is in line with the overall batch size used for training.

For the SPMat SimCLR model, we used a batch size of 256 to include more negative samples in a mini-batch. The loss function used a temperature parameter of 0.03 to control the contrastive loss scale, with the contrast mode set to 'all.' A base temperature of 0.03 was also used to further refine the contrastive learning process. 

For fine-tuning the model, we used a randomly initialized MLP head with two fully connected layers for property prediction purposes. In addition, we employ a batch size of 128 and train for 200 epochs, with performance evaluations conducted at the end of each epoch. The fine-tuning process started from a pre-trained checkpoint while model performance was logged every 50 steps to monitor progress and make necessary adjustments during training. The fine-tuning task was set up as a regression problem. We used the Adam optimizer with a learning rate of 0.001 and a momentum of 0.9. For data loading, a validation split of 10\%  was applied to monitor the model performance on unseen data, while a test split of 20\%  was reserved for final evaluation. This setup ensured that the model was effectively fine-tuned to predict material properties in a regression context. 

We generated the t-SNE plots for both the pretrained and finetuned models. In all the cases the parameters to generate t-SNE were kept same for example the batch size was kept 8, model and dataset parameters similar to the trained model. Only test data set has been used to generate the embedding plots for fair judgment. 

%\section*{Barlow-Twin Extension}
% \noindent LaTeX formats citations and references automatically using the bibliography records in your .bib file, which you can edit via the project menu. Use the cite command for an inline citation, e.g.  \cite{Hao:gidmaps:2014}.

% For data citations of datasets uploaded to e.g. \emph{figshare}, please use the \verb|howpublished| option in the bib entry to specify the platform and the link, as in the \verb|Hao:gidmaps:2014| example in the sample bibliography file.
\section*{Data Availability}
All crystallographic information files (cif) for recreating different materials and the required labels for pretraining and fine-tuning are available publicly and can be accessed using the Materials Project website (\url{https://materialsproject.org/}) and their api. 

\section*{Code Availability}
The code developed for this study will be shared on a public GitHub repository after the work is accepted.

\section*{Acknowledgements}
We thank the support of the West Virginia High Education Policy Commission under the call Research Challenge Grand Program 2022, RCG 23-007 Award.
We also thank the Pittsburgh Supercomputer Center (Bridges2) and the San Diego Supercomputer Center (Expanse) through the allocation DMR140031 from the Advanced Cyberinfrastructure Coordination Ecosystem: Services \& Support (ACCESS) program, which National Science Foundation supports grants \#2138259, \#2138286, \#2138307, \#2137603, and \#2138296. 
We also recognize the computational resources provided by the WVU Research Computing Dolly Sods HPC cluster, which is funded in part by NSF OAC-2117575.
We also recognize the support of the NASA EPSCoR Award 80NSSC22M0173.

\section*{Author contributions statement}

% Must include all authors, identified by initials, for example:
% A.A. conceived the experiment(s),  A.A. and B.A. conducted the experiment(s), C.A. and D.A. analysed the results.  All authors reviewed the manuscript. 
% C.M.A.R. Conceptualization, Data curation, Investigation, Formal analysis, Methodology, Software,
% Visualization, Writing–original draft, Writing–review and editing. K.C. Methodology, Software, Validation, Writing–review and editing. AHR: Formal analysis, Funding acquisition, Methodology, Writing–review and editing P.K.G. Funding acquisition, Project administration, Resources, Supervision, Validation, Writing–review and editing.
C.M.A.R. conceived the study. C.M.A.R. and P.K.G. developed the methodology and software. C.M.A.R., and A.H.R. conducted the formal analysis. C.M.A.R. prepared the original draft, and A.H.R., and P.K.G. reviewed and edited the manuscript. P.K.G. supervised the project and secured funding, along with A.H.R.. All authors contributed to the final manuscript.

\section*{Competing interests}
All authors declare that they have no financial or non-financial competing interests. 

% \section*{Data availability}
% All data utilized in this study are publicly accessible. The authors sourced datasets from the Materials Project, a publicly available data repository. These datasets are available on the Materials Project website (https://materialsproject.org/).

% \section*{Additional information}

% To include, in this order: \textbf{Accession codes} (where applicable); \textbf{Competing interests} (mandatory statement). 

% The corresponding author is responsible for submitting a \href{http://www.nature.com/srep/policies/index.html#competing}{competing interests statement} on behalf of all paper authors. This statement must be included in the submitted article file.

% \begin{figure}[ht]
% \centering
% \includegraphics[width=\linewidth]{stream}
% \caption{Legend (350 words max). Example legend text.}
% \label{fig:stream}
% \end{figure}

% \begin{table}[ht]
% \centering
% \begin{tabular}{|l|l|l|}
% \hline
% Condition & n & p \\
% \hline
% A & 5 & 0.1 \\
% \hline
% B & 10 & 0.01 \\
% \hline
% \end{tabular}
% \caption{\label{tab:example}Legend (350 words max). Example legend text.}
% \end{table}

% Figures and tables can be referenced in LaTeX using the ref command, e.g. Figure \ref{fig:stream} and Table \ref{tab:example}.
\bibliography{sample}

\end{document}